\pdfoutput=1

\documentclass[11pt]{article}

\usepackage[preprint]{acl}

\usepackage{times}
\usepackage{latexsym}

\usepackage[T1]{fontenc}

\usepackage[utf8]{inputenc}

\usepackage{microtype}

\usepackage{inconsolata}

\usepackage{graphicx}
\usepackage{hyperref}
\usepackage{url}
\usepackage{comment}
\usepackage{booktabs}
\usepackage{subcaption}
\usepackage{verbatimbox}
\usepackage{microtype}
\usepackage{multirow}
\usepackage{longtable} 
\usepackage{bbm}
\usepackage{mathtools}
\usepackage{amssymb}
\usepackage{pifont}
\title{Knowledge-enhanced Multimodal ECG Representation Learning with Arbitrary-Lead Inputs}

\author{
Che Liu\textsuperscript{1}, 
Cheng Ouyang\textsuperscript{2}, 
Zhongwei Wan\textsuperscript{3}, 
Haozhe Wang\textsuperscript{4}, 
\bf Wenjia Bai\textsuperscript{1},  
\bf Rossella Arcucci\textsuperscript{1} \\
\textsuperscript{1}Imperial College London, UK,
\textsuperscript{2}University of Oxford, UK,
\textsuperscript{3}Ohio State University, US, \\
\textsuperscript{4}Hong Kong University of Science and Technology, China\\
\href{mailto:che.liu21@imperial.ac.uk}{che.liu21@imperial.ac.uk} \\
}

\begin{document}
\maketitle
\begin{abstract}
Recent advances in multimodal ECG representation learning center on aligning ECG signals with paired free-text reports. However, suboptimal alignment persists due to the complexity of medical language and the reliance on a full 12-lead setup, which is often unavailable in under-resourced settings. To tackle these issues, we propose \textbf{K-MERL}, a knowledge-enhanced multimodal ECG representation learning framework. \textbf{K-MERL} leverages large language models to extract structured knowledge from free-text reports and employs a lead-aware ECG encoder with dynamic lead masking to accommodate arbitrary lead inputs. Evaluations on six external ECG datasets show that \textbf{K-MERL} achieves state-of-the-art performance in zero-shot classification and linear probing tasks, while delivering an average \textbf{16\%} AUC improvement over existing methods in partial-lead zero-shot classification\footnote{All data and code will be released upon acceptance.}.
\end{abstract}

\section{Introduction}
\label{sec: intro}
Recent advancements in deep learning have enabled automated classification of cardiovascular disease (CVD) using electrocardiograms (ECGs), one of the most crucial diagnostic tools. However, most methods are supervised, requiring large amounts of annotated data, which is costly and demands prohibitively extensive expert effort in annotation \citep{scdnn,SPNv2}.
To address this challenge, self-supervised multimodal learning has recently emerged as an effective approach for learning representative ECG features from accompanied free-text clinical reports~\citep{li2023frozen,pham2024c,merl}. 
To this end, MERL~\citep{merl} recently introduced the first comprehensive benchmark using the largest dataset MIMIC-ECG~\citep{mimicecg} for pretraining, and six datasets~\citep{wagner2020ptb,cpsc2018,csn1,csn2} for evaluating downstream task performance, including zero-shot classification and linear probing. 

Despite outperforming signal-only self-supervised approaches, multi-modal approaches, including MERL \citep{merl}, still have notable drawbacks: They directly align ECG signals with reports, introducing unnecessary noise due to the free-text nature of the reports, and failing to fully exploit the rich cardiac knowledge contained within the text. 
Additionally, they encode ECG in a lead-agnostic manner, overlooking the unique spatial and temporal characteristics of the individual 12 ECG leads. Moreover, they require all 12 leads to be available as input, limiting their ability to generalize across different lead combinations. 
This raises important practical concerns since full 12-lead ECG data is not always available in clinical environments due to factors such as patient mobility issues, the need for rapid assessments in emergencies, and limited resource in pre-hospital care environments \citep{bray2021single,swor2006prehospital,quinn2020has,nonogi2008abstract,kotelnik202112,zhang2019all,nonogi2008abstract}.

To overcome the challenges listed above, we make the following contributions:  
\textbf{(1)} We propose a framework dubbed \textbf{K}nowledge-enhanced \textbf{E}CG \textbf{M}ultimodal \textbf{R}epresentation \textbf{L}earning (\textbf{K-MERL}), which extracts cardiac-related entities from free-text ECG reports, converting unstructured reports into structured knowledge to enhance self-supervised ECG multimodal learning. To the best of our knowledge, this is the first work to leverage structured cardiac entities extracted from clinical reports to improve ECG multimodal learning.
\textbf{(2)} To effectively capture and leverage the lead-specific spatial and temporal characteristics of 12-lead ECGs, we explore various tokenization and positional embedding techniques. In particular, we design \textit{lead-specific tokenization} and \textit{lead-specific spatial positional embeddings}, enabling the framework to capture the distinctiveness of each lead.
\textbf{(3)} To enable our framework to handle arbitrary combinations of input leads, we introduce a \textit{dynamic lead masking} strategy. In addition, we propose an \textit{independent segment masking} strategy to further capture lead-specific temporal patterns.
\textbf{(4)} Our K-MERL framework demonstrates superior performance in zero-shot classification and linear probing on multiple downstream datasets in various lead combinations, from a single lead to all 12 leads.

\begin{figure*}[t!]
    \centering
    \vspace{-5pt}
    \includegraphics[width=0.95\linewidth]{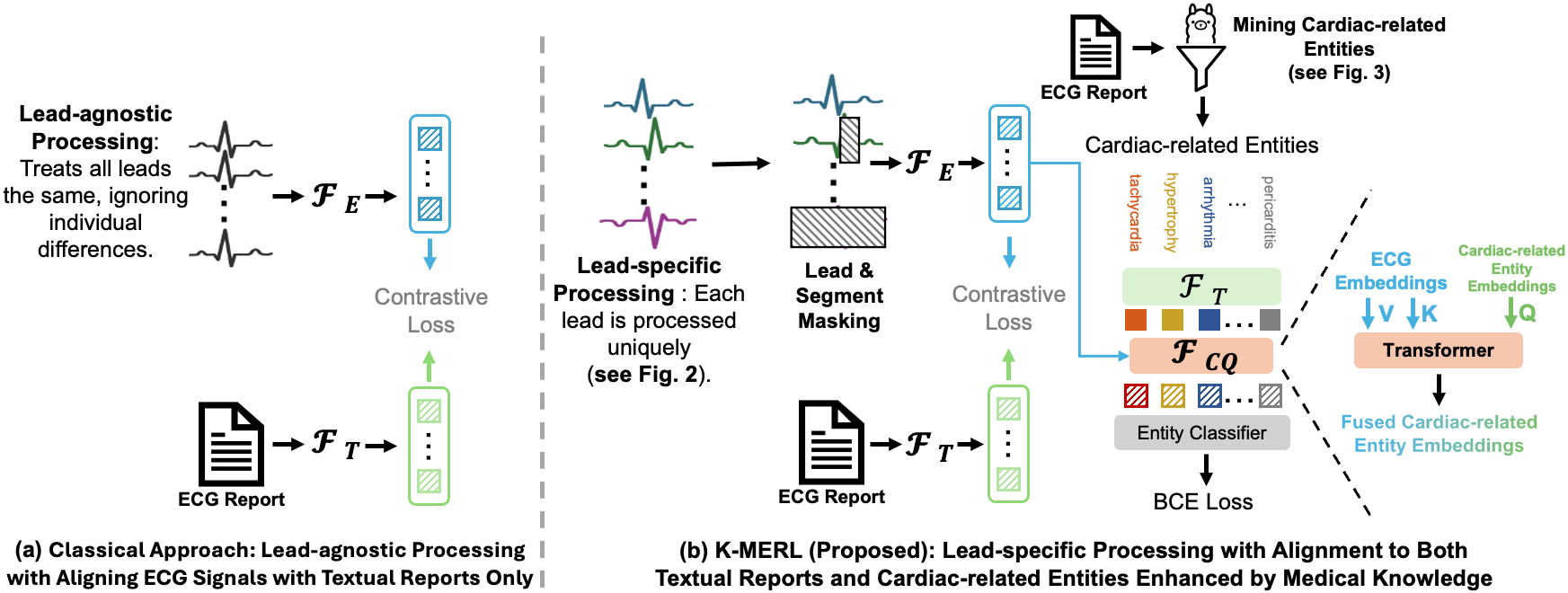}
    \vspace{-8pt}
    \caption{
    Comparison between classical ECG multimodal learning and our K-MERL framework.  
\textbf{(a):} The classical approaches (e.g., MERL \citep{merl}) are suboptimal: they processes all leads in a lead-agnostic manner and naively align ECG signals directly free-text reports.  
\textbf{(b):} K-MERL introduces lead-specific processing and lead \& segment masking to capture spatial-temporal patterns unique to each lead. It also extracts cardiac-related entities from reports as structured knowledge and aligns them with ECG features to enhance multimodal learning, thereby reducing the complexity introduced by the grammatical structure of free-text reports.
    }
    \label{fig: framework}
    \vspace{-15pt}
\end{figure*}

\section{Method}
\label{sec: method}

\subsection{Overview}
To this end, we first utilize a general-purpose \textit{open-source} large language model (LLM), such as Llama3.1 \citep{llama3modelcard}, without domain-specific fine-tuning, to extract cardiac-related entities from free-text ECG reports.\footnote{Entity extraction is inherently simpler than high-level text comprehension in specialized domains, and has been shown effective with general-purpose LLMs \citep{kad}.} This makes our approach adaptable and well-positioned to benefit from future advancements in LLMs. Additionally, we design a lead-aware ECG encoder with \textit{lead and segment masking} strategies, allowing the model to handle arbitrary lead inputs while capturing lead-specific spatial-temporal patterns.

Our overall framework is illustrated in Fig \ref{fig: framework}(b), shown together with the previous state-of-the-art MERL that is based on naive cross-modal contrastive learning \citep{merl}, in Fig \ref{fig: framework}(a). While both approaches utilize contrastive learning with an ECG signal encoder $\mathcal{F}_\textrm{E}$ processing signal inputs and a text encoder $\mathcal{F}_\textrm{T}$ processing reports, our method introduces substantial innovations, including lead-specific processing, dynamic masking strategies, and the extraction of cardiac-related entities from free-text reports, significantly enhancing ECG multimodal learning. 

In the following sections, we introduce the model framework and lead-specific processing in Sec \ref{sec: framework}, followed by the proposed masking strategies in Sec \ref{sec: masking}. We then describe the pipeline for extracting cardiac-related entities as structured knowledge from ECG reports in Sec \ref{sec: mining condition}. Finally, in Sec \ref{sec: training obj}, we explain the knowledge-enhanced ECG multimodal learning process, a synergy of the aforementioned components.

\subsection{Lead-specific Processing}
\label{sec: framework}
\begin{figure*}[t!]
    \centering
    \includegraphics[width=0.99\textwidth]{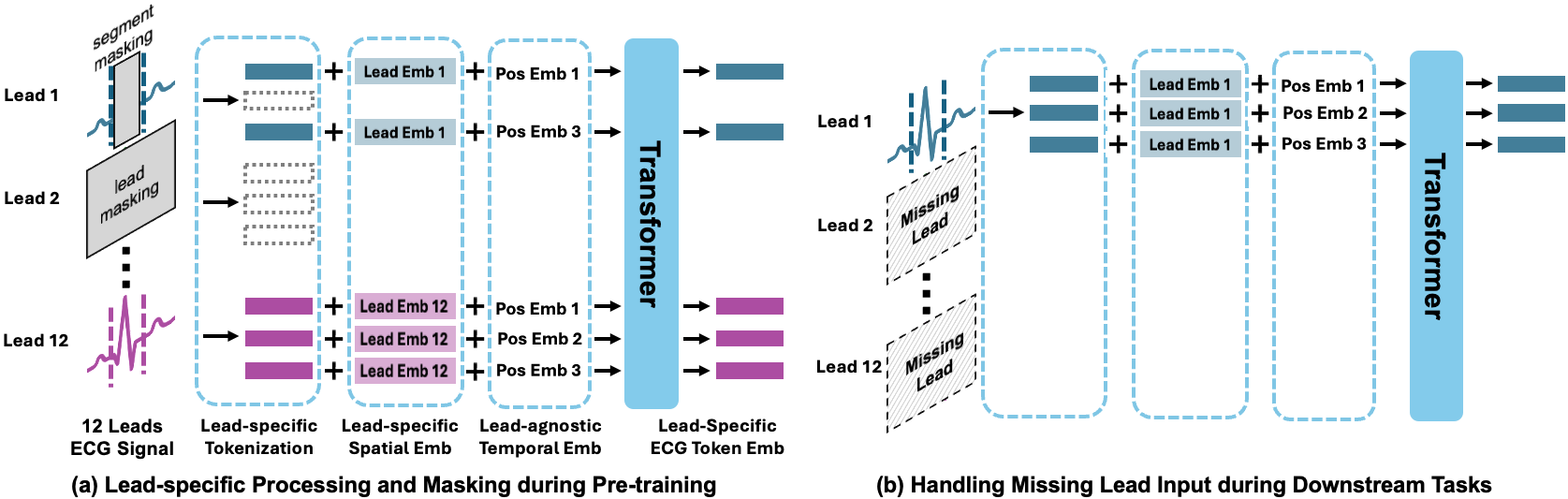}
    \caption{
\small{Illustration of our lead-specific processing and handling of partial leads input in K-MERL.
\textbf{(a):} Lead-specific processing and masking during pre-training. The model employs lead-specific tokenization, spatial embeddings, and lead-agnostic temporal embeddings to capture spatial-temporal patterns for each lead (see Sec \ref{sec: framework}). Dynamic lead masking is used to simulate inputs with arbitrary combinations of leads, while segment masking encourage the framework to captures temporal patterns (see Sec \ref{sec: masking}).
\textbf{(b):} Handling partial lead input during downstream tasks. When leads are missing, the model processes only the available leads using lead-specific embeddings, allowing maintained performance even with incomplete data.}
    }
    \label{fig: encoder}
    \vspace{-5pt}
\end{figure*}

To begin with, we define the symbols used in our framework: Given a training dataset $\mathcal{X}$ consisting of $N$ ECG-report pairs, we represent each pair as $(\mathbf{e}_{i}^{l}, \mathbf{t}_{i})$, where $\mathbf{e}_{i}^{l} \in \mathcal{E}$ denotes the raw 12-lead ECG signals for lead $l \in \{1, 2, 3, \dots, 12\}$ of the $i$-th subject ($i = 1, 2, 3, \dots, N$), and $\mathbf{t}_{i} \in \mathcal{T}$ represents the associated free-text report. We then perform lead-specific processing, as illustrated in Fig \ref{fig: encoder}.

\noindent \textbf{Lead-specific Tokenization.}  
Consider an input ECG signal \( e_i^l \) with 12 leads and a signal length denoted by \( S \). We split the time-series signal into $M$ non-overlapping segments, each segment of length $\frac{S}{M}$, and perform tokenization for them. In this way, each lead ECG is projected into a sequence of tokens:
\vspace{-1em}
\begin{equation*}
e_{i}^{l}\left[p_1\right], e_{i}^{l}\left[p_2\right], e_{i}^{l}\left[p_3\right], \ldots, e_{i}^{l}\left[p_M\right]
\end{equation*}
\vspace{-2em}

where \( e_{i}^{l}\left[p_m\right] \) corresponds to the ECG token for the \( m \)-th segment for lead \( l \). For 12 leads, the total number of tokens is $12 \times M$. Unlike MERL \citep{merl}, which generates a single token for a 12-lead ECG temporal segment, we produce tokens separately for each individual lead to capture the lead-specific nature.

\noindent \textbf{Lead-specific Spatial Positional Embedding.}  
We apply a learnable linear projection \( \mathbf{W} \in \mathbb{R}^{p \times d} \) to each token \( e_{i}^{l}\left[p_m\right] \). Then, we introduce learnable \textit{lead embeddings} \([\texttt{lead}_1, \ldots, \texttt{lead}_{12}]\), where \(\texttt{lead}_l \in \mathbb{R}^d\), to capture the characteristics of each lead. The resulting input sequence can be written as:
\vspace{-0.7em}
\begin{align*}
    &\texttt{lead}_1 +  \mathbf{W}e_i^l[p_1], \ldots , 
    \texttt{lead}_1 +  \mathbf{W}e_i^l[p_M], \ldots, \\
    &\texttt{lead}_{12} +  \mathbf{W}e_i^l[p_1], \ldots , 
    \texttt{lead}_{12} +  \mathbf{W}e_i^l[p_M].
\end{align*}
\vspace{-1.7em}

\noindent \textbf{Lead-agnostic Temporal Positional Embedding.}  
In line with lead-specific spatial positional embedding, we also incorporate \textit{learnable lead-agnostic temporal embeddings} to retain the temporal information of ECG signals. These embeddings are denoted as \([\texttt{temp}_1, \ldots, \texttt{temp}_M]\), where \(\texttt{temp}_m \in \mathbb{R}^d\). It is worth noting that these positional embeddings are shared across leads, enabling the model to recognize temporal properties across leads, as all leads originate from the same source and share the same temporal domain properties. The resulting input sequence can be written as:
\vspace{-0.7em}
\begin{align*}
    &\texttt{temp}_1 + \texttt{lead}_1 + We_i^l[p_1], \\
    & \quad \ldots \ , 
    \quad \texttt{temp}_M + \texttt{lead}_1 + We_i^l[p_M], \\
    &\quad \ldots , \quad 
    \texttt{temp}_{1} + \texttt{lead}_{12} + We_i^l[p_1], \\
    &\quad \ldots \ , 
    \quad \texttt{temp}_{M} + \texttt{lead}_{12} + We_i^l[p_M].
\end{align*}
\vspace{-2em}

\subsection{Lead and Segment Masking}
\label{sec: masking}
Using a fixed number of masked leads limits the model's flexibility in handling arbitrary lead inputs. To address this, we propose \textbf{D}ynamic \textbf{L}ead \textbf{M}asking \textbf{(DLM)}, enabling the model to handle varying lead combinations (Fig. \ref{fig: encoder} a). For an ECG signal \( e_{i}^{l} \) with 12 leads, we first randomly sample a number from \(\{9, 10, 11\}\), which determines how many leads will be \textit{masked}. Then, we randomly select a set of unmasked lead indices, denoted as \( \hat{l} \), and mask the remaining leads. This approach ensures the model is exposed to diverse combinations of unmasked and masked leads during pretraining. The resulting ECG signal with the selected unmasked leads is denoted as \( e_{i}^{\hat{l}} \).

To better capture the temporal patterns of each ECG lead, we introduce \textbf{L}ead-independent \textbf{S}egment \textbf{M}asking \textbf{(LSM)} (Fig. \ref{fig: encoder} a). Applying masking across all tokens from an ECG signal could lead to imbalances, where some leads have more masked tokens than others. To avoid this, LSM applies masking separately to each lead, ensuring an equal number of masked tokens per lead. For each unmasked lead signal \( e_{i}^{\hat{l}} \), we randomly select masked token indices \( \mathcal{H}^{\hat{l}} \) based on a masking proportion of 0.25. The model then processes only the unmasked tokens, denoted as \( \{ e_{i}^{\hat{l}}[p_h] \}_{h \notin \mathcal{H}^{\hat{l}}} \).

In the experiments we ablate DLM or LSM to verify their effectiveness, as shown in Tab \ref{tab: abla mask strategy} and Fig \ref{fig: abla token mask}.

\subsection{Mining Cardiac-related Entities from Report}
\label{sec: mining condition}

\begin{figure*}[t!]
    \centering
    \scalebox{0.9}{
    \includegraphics[width=0.99\linewidth]{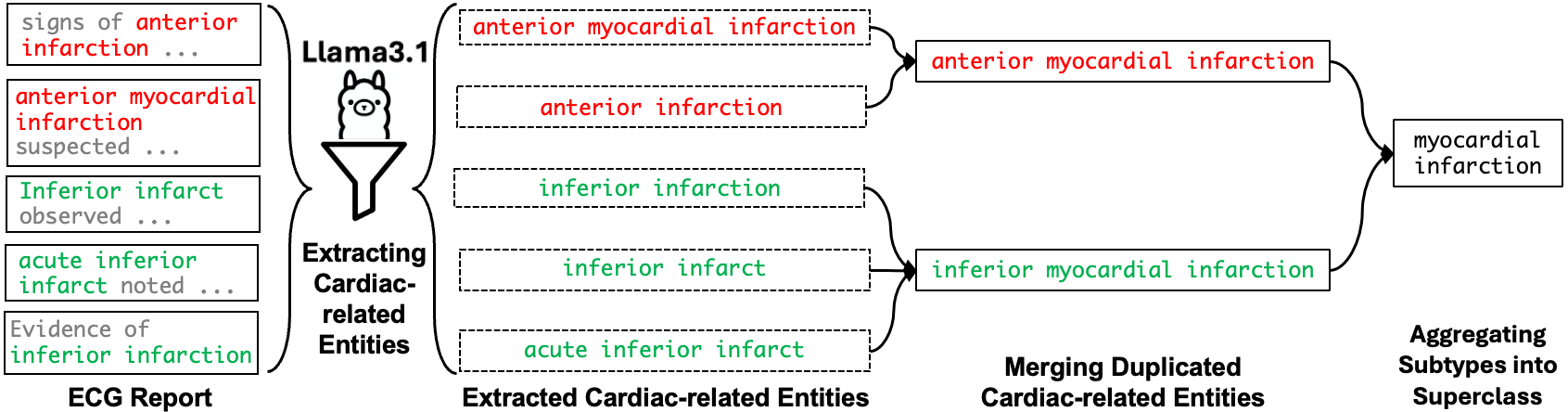}
     }
    \caption{
    Illustration of mining structured knowledge from free-text reports (see Sec \ref{sec: mining condition}). First, cardiac-related entities are extracted from free-text ECG reports using an open-source LLM (e.g., Llama3.1-70B-Instruct). Next, we query the LLM to merge duplicated or synonymous cardiac-related entities into a list of unique names. Finally, the LLM detects and aggregates subtypes into their respective superclasses, creating a structured hierarchy of cardiac-related entities.
    }
    \label{fig: mining condition}
    \vspace{-15pt}
\end{figure*}

In this section, we introduce the structured knowledge extraction process for handling free-text ECG reports. The pipeline is illustrated in Fig \ref{fig: mining condition}. Since each ECG report provides descriptions of cardiac-related entities, as shown in the leftmost part of Fig. \ref{fig: mining condition}, our goal is to extract all positive cardiac-related entities mentioned in the report as structured knowledge to enhance the supervision signals for ECG multimodal learning.

\noindent\textbf{Extracting Cardiac-related Entities.}
Unlike existing biomedical multimodal learning approaches from the radiology domain, which rely on knowledge graphs to extract structured knowledge from reports \citep{kad,wu2023medklip}, we directly query an LLM with the following prompt: `\texttt{Please extract all positive Cardiac-related Entities from the given ECG report. Output format is [Entity1, Entity2, ...]}'. 
There are two main reasons for this approach. First, there is no off-the-shelf knowledge graph (KG) specifically focused on ECG, making it impractical to use KG-based methods for extracting structured knowledge. Second, since we are only extracting existing terms from the free-text report, we can easily verify that the extracted cardiac-related entities are present and positive, ensuring no non-existent terms are generated by the LLM. Moreover, \citep{kad} has already demonstrated that a general-purpose LLM can effectively extract existing medical terms from free-text reports independently of any external knowledge database.
To ensure accuracy, after each extraction operation, we query the LLM with: `\texttt{Please verify the extracted cardiac-related entities as existing and positive in the given report. Output format is YES or NO}', and only retain the cardiac-related entities with a \texttt{`YES'} response. After this stage, we obtain a total of 341 unique cardiac-related entities in the whole dataset..

\noindent\textbf{Merging Duplicated Cardiac-related Entities.}
After extracting all cardiac-related entities from whole dataset, we observe that many names share the same semantics but are expressed differently, as shown in the second part of Fig \ref{fig: mining condition}. This variation arises because different clinical protocols generate ECG reports in different styles, even though they describe the same cardiac-related entities. To address this, we query the LLM with: `\texttt{Please merge the cardiac-related entities that have the same semantics but different expressions. Here are <all Cardiac-related Entities>. Output format is JSON, where the key is the original name and the value is the merged name.}'
 After this stage, we obtain a total of 252 unique cardiac-related entities in the whole dataset..
 
\noindent\textbf{Aggregating Subtypes into Superclasses.}
Since cardiac-related entities are organized in a clear hierarchical structure \citep{arnaout2016peripartum,okshina2019p5352}, for example, as shown in the rightmost part of Fig \ref{fig: mining condition}, \texttt{`anterior myocardial infarction'} and \texttt{`inferior myocardial infarction'} are subtypes of the superclass \texttt{`Myocardial infarction'} \citep{brieger2000hierarchy}, we query the LLM with the following prompt: `\texttt{Please detect all the superclasses present in <all Cardiac-related Entities>. Output format is JSON, where the key is the superclass name and the values are the cardiac-related entities that belong to this superclass.}'

After this stage, we identify 25 superclasses of cardiac-related entities. By the end of the process, we obtain a list of 277 unique cardiac-related entities for the entire dataset. The list of these entities is represented as \( \mathcal{Q} = \{q_1, q_2, \ldots, q_Q\} \), where \( Q = 277 \).
 For each ECG report $\textbf{t}_i$, we create a label vector of length 277, where the positions corresponding to present and positive cardiac-related entity are set to 1, and all other positions are set to 0. This results in a binary label vector for each report, which we denote as \( \mathbf{y}_i \in \{0, 1\}^{277} \).

\subsection{Knowledge-enhanced ECG Multimodal Learning}
\label{sec: training obj}

\noindent\textbf{Aligning ECG and Reports.}
In this framework, as shown in Fig \ref{fig: framework} (b), two distinct encoders for ECG signals and text reports, symbolized as \( \mathcal{F}_{\textrm{E}} \) and \( \mathcal{F}_{\textrm{T}} \), transform the sample pair \(( \mathbf{e}_{i}, \mathbf{t}_{i} ) \) into the latent embedding space, represented as \( (\mathbf{z}_{e,i},  \mathbf{z}_{t,i} ) \).
The dataset at the feature level is then denoted as \(\mathcal{X}=\left\{\left(\mathbf{z}_{e,1}, \mathbf{z}_{t,1}\right),\left(\mathbf{z}_{e,2}, \mathbf{z}_{t,2}\right), \ldots, \left(\mathbf{z}_{e,N}, \mathbf{z}_{t,N}\right)\right\}\), where \(\mathbf{z}_{e,i} = \mathcal{F}_{\textrm{E}}(\mathbf{e}_{i})\) and \(\mathbf{z}_{t,i} = \mathcal{F}_{\textrm{T}}(\mathbf{t}_{i})\).
Afterward, two non-linear projectors for ECG and text embeddings, denoted as \( \mathcal{P}_{e} \) and \( \mathcal{P}_{t} \), transform \( \mathbf{z}_{e,i} \) and \( \mathbf{z}_{t,i} \) into the same dimensionality \( d \), with \(\mathbf{\hat{z}}_{e,i} = \mathcal{P}_{e}(\mathrm{AvgPool}(\mathbf{z}_{e,i}))\) and \(\mathbf{\hat{z}}_{t,i} = \mathcal{P}_{t}(\mathrm{AvgPool}(\mathbf{z}_{t,i}))\).
Next, we compute the cosine similarities as \(s_{i,i}^{e2t} = \mathbf{\hat{z}}_{e,i}^{\top} \mathbf{\hat{z}}_{t,i}\), representing the ECG-report similarities, and formulate the ECG-report contrastive loss $\mathcal{L}_{\mathrm{contrast}}$.

\begin{align}
&\mathcal{L}^{e2t}_{i,j} = -\log \frac{\exp(s_{i,j}^{e2t} / \tau)}{\sum_{k=1}^{L}{\mathbbm{1}_{[k \neq i]}\exp(s_{i,k}^{e2t} /\eta)}},  \notag \\
&\mathcal{L}_{\mathrm{contrast}} = \frac{1}{L} \sum_{i=1}^N\sum_{j=1}^N\mathcal{L}^{e2t}_{i,j}.
\end{align}
\vspace{-1.5em}

The temperature hyper-parameter, denoted as \(\eta\), is set to 0.07 in our study. \(L\) refers to the batch size per training step, which is a subset of \(N\).

\noindent\textbf{Aligning ECG and Cardiac-related Entities.}
To learn the knowledge from extracted cardiac-related entities, we design a cardiac query network, denoted as \(\mathcal{F}_{\textrm{CQ}}\). This network consists of four transformer layers concatenated with a linear classifier that predicts each ECG's corresponding cardiac entity labels \(\mathbf{y}_i\).
Given the set of cardiac-related entities \(\mathcal{Q}\), we compute a corresponding set of cardiac query vectors using the text encoder, denoted as \(\mathcal{\mathbf{Q}}=\left\{\mathbf{q}_1, \mathbf{q}_2, \ldots, \mathbf{q}_Q\right\}\), where each query vector \(\mathbf{q}_i\) is obtained as \(\mathbf{q}_i = \mathcal{F}_{\mathrm{T}}(q_i)\). These query vectors are then used as inputs for the cardiac query network \(\mathcal{F}_\mathrm{CQ}\).
During pre-training, the ECG features \(\mathbf{z}_{e,i}\) serve as the key and value inputs to the cardiac query network \(\mathcal{F}_\mathrm{CQ}\).
We use binary cross-entropy (BCE) loss to compute the predictions from \(\mathcal{F}_{\textrm{CQ}}\) and compare them to the existence labels \(\mathbf{y}_i\). The total loss is defined as:
\vspace{-1em}
\begin{align}
    &\mathcal{L}_{\mathrm{CQ}} = \frac{1}{L} \sum_{i=1}^N \textrm{BCE} (\mathcal{F}_\mathrm{CQ}(\mathcal{\mathbf{Q}}, \mathcal{\mathbf{z}}_{e,i}), \mathbf{y}_i), \notag \\
    &\mathcal{L}_{\mathrm{total}} = \mathcal{L}_{\mathrm{contrast}} + \mathcal{L}_{\mathrm{CQ}}.
\end{align}
\vspace{-2.5em}

\section{Experiments}
\label{sec: exp}
\subsection{Pre-training Configurations}
\noindent \textbf{MIMIC-ECG.} 
We pre-train K-MERL using the MIMIC-ECG dataset \citep{mimicecg}, comprising 800,035 ECG-report pairs. Each sample includes a raw ECG signal recorded at 500Hz over a 10-second duration, along with its corresponding report. For fair comparison with the MERL framework \citep{merl}, we adhere to their preprocessing protocol, available in the official GitHub repository\footnote{https://github.com/cheliu-computation/MERL-ICML2024/tree/main}. After preprocessing, we obtain 771,693 samples for model pre-training.
\\
\noindent \textbf{Implementation.}
For pre-training, we inherit the settings from MERL \citep{merl}, using a ViT-tiny model as the ECG encoder and Med-CPT \citep{jin2023medcpt} as the text encoder. The key differences in our approach are the proposed lead-specific tokenizer and  spatial-temporal positional embeddings. For extracting cardiac-related entities from the ECG reports, we utilize Llama3.1-70B-Instruct \citep{llama3modelcard}, with ablations of different LLMs shown in Tab \ref{tab:abla_llm}. Pre-training configuration details are provided in Sec \ref{sec: pre config}.

\subsection{Downstream Tasks Configurations}
We evaluate our framework on both zero-shot classification and linear probing, using full and partial lead ECGs across multiple public datasets covering over 100 cardiac conditions. We adhere to the data split and preprocessing provided by MERL \citep{merl}. The tasks are implemented on the following datasets:
\textbf{(1) PTBXL:} The PTBXL dataset \citep{wagner2020ptb} includes 21,837 ECG signals from 18,885 patients, sampled at 500 Hz for 10 seconds. It provides four subsets for multi-label classification: \textbf{Superclass} (5 categories), \textbf{Subclass} (23 categories), \textbf{Form} (19 categories), and \textbf{Rhythm} (12 categories), with varying sample sizes.
\textbf{(2) CPSC2018:} The CPSC2018 dataset \citep{cpsc2018} contains 6,877 12-lead ECG records, sampled at 500 Hz, annotated with 9 distinct labels.
\textbf{(3) CSN:} The Chapman-Shaoxing-Ningbo (CSN) dataset \citep{csn2,csn1} comprises 45,152 ECG records sampled at 500 Hz for 10 seconds. After excluding records with `unknown' annotations, the final curated dataset includes 23,026 ECG records with 38 labels.
Detailed information about the downstream datasets is presented in Tab \ref{tab:split}.

In the downstream tasks, we implement three scenarios: zero-shot classification, linear probing, and partial lead analysis. The implementation details are provided in Sec \ref{sec: downstream implment}.

\subsection{State-of-the-art on Zero-shot Classification}
\label{sec: zeroshot cls}
\begin{figure*}[t!]
    \centering
    \parbox[h]{.63\textwidth}{
    \includegraphics[width=1.0\linewidth]{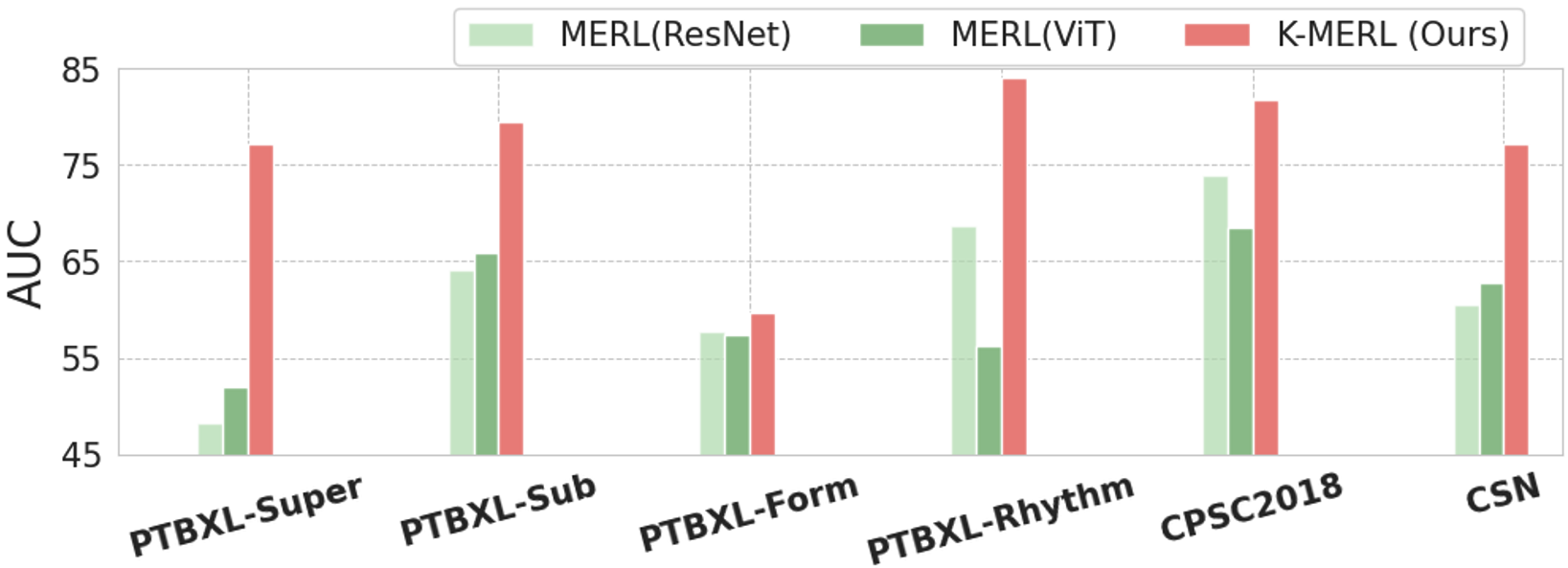}
    }
    \hspace{\fill}
    \parbox[h]{.35\textwidth}{
    \caption{
        \small{Performance on zero-shot classification across six datasets, comparing K-MERL with previous ECG multimodal learning methods. Notably, we use the original disease category names as prompts for both K-MERL and MERL to ensure a fair comparison. 
        }
    }
    \label{fig: zeroshotres}
    }
    \vspace{-10pt}
\end{figure*}

We first evaluate K-MERL on zero-shot classification using 12-lead input across all downstream datasets. The results for each dataset, along with the average AUC score across six datasets, are shown in Fig \ref{fig: zeroshotres}. Our framework significantly outperforms MERL with both backbone architectures, demonstrating the superiority of K-MERL when using the original disease names as text prompts. 



\begin{figure*}[t!]
    \centering
    \parbox[h]{.63\textwidth}{
    \includegraphics[width=1.\linewidth]{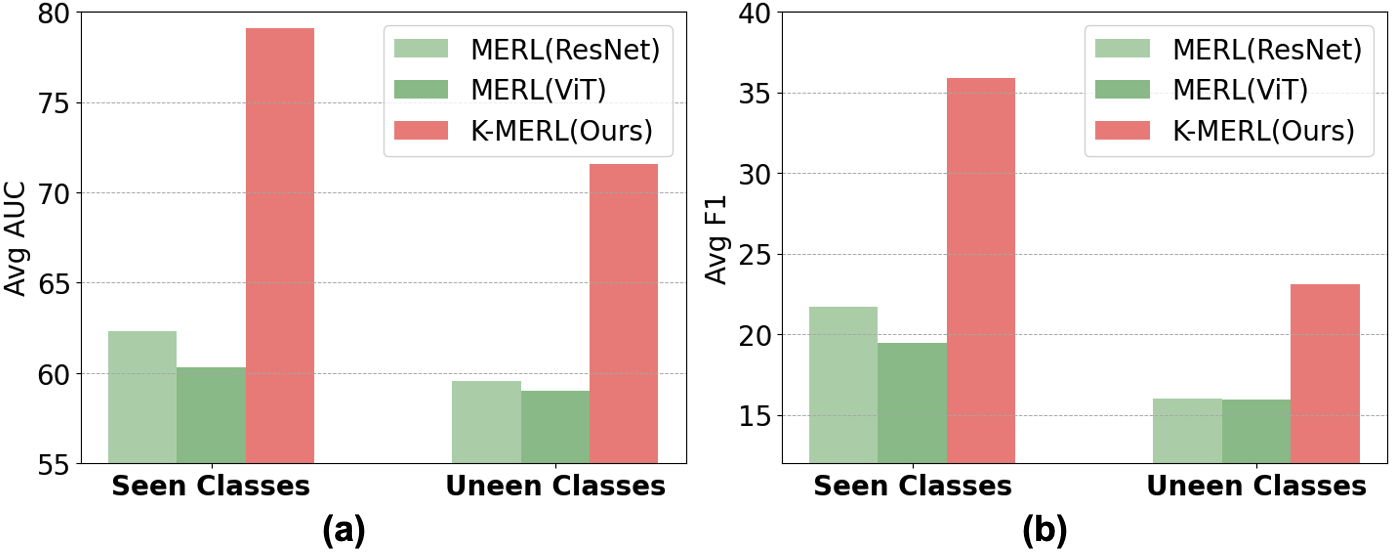}
    }
    \hspace{\fill}
    \parbox[h]{.35\textwidth}{
    \caption{
        \small{Comparison of K-MERL and MERL on seen and unseen classes, reporting \textbf{(a)} Average AUC and \textbf{(b)} Average F1 scores. Definitions are in Sec \ref{sec: zeroshot cls}.
        }
    }
    \label{fig: seen_unseen_new}
    }
    \vspace{-10pt}
\end{figure*}


\begin{table*}[t!]
\centering
\caption{\small{Linear probing results of K-MERL and other ECG learning methods, with best results \textbf{bolded}.}}
\vspace{-8pt}
\scalebox{0.52}{    
\begin{tabular}{lccc|ccc|ccc|ccc|ccc|ccc}
    \toprule[1.2pt]
     & \multicolumn{3}{c}{PTBXL-Super} & \multicolumn{3}{c}{PTBXL-Sub} 
     & \multicolumn{3}{c}{PTBXL-Form} & \multicolumn{3}{c}{PTBXL-Rhythm} 
     & \multicolumn{3}{c}{CPSC2018} & \multicolumn{3}{c}{CSN} \\
    Method & 1\% & 10\% & 100\% & 1\% & 10\% & 100\% & 1\% & 10\% & 100\% 
           & 1\% & 10\% & 100\% & 1\% & 10\% & 100\% & 1\% & 10\% & 100\% \\
    \midrule[1.2pt]
    \multicolumn{19}{l}{\textbf{\textit{From Scratch}}}\\
    \midrule
    Random Init (CNN) & 70.45 & 77.09 & 81.61 & 55.82 & 67.60 & 77.91 & 55.82 & 62.54 & 73.00 & 46.26 & 62.36 & 79.29 & 54.96 & 71.47 & 78.33 & 47.22 & 63.17 & 73.13 \\
    Random Init (Transformer) & 70.31 & 75.27 & 77.54 & 53.36 & 67.56 & 77.43 & 53.47 & 61.84 & 72.08 & 45.36 & 60.33 & 77.26 & 52.93 & 68.0 & 77.44 & 45.55 & 60.23 & 71.37 \\
    \midrule
    \multicolumn{19}{l}{\textbf{\textit{ECG only SSL}}}\\
    \midrule
    SimCLR & 63.41 & 69.77 & 73.53 & 60.84 & 68.27 & 73.39 & 54.98 & 56.97 & 62.52 & 51.41 & 69.44 & 77.73 & 59.78 & 68.52 & 76.54 & 59.02 & 67.26 & 73.20 \\
    BYOL & 71.70 & 73.83 & 76.45 & 57.16 & 67.44 & 71.64 & 48.73 & 61.63 & 70.82 & 41.99 & 74.40 & 77.17 & 60.88 & 74.42 & 78.75 & 54.20 & 71.92 & 74.69 \\
    BarlowTwins & 72.87 & 75.96 & 78.41 & 62.57 & 70.84 & 74.34 & 52.12 & 60.39 & 66.14 & 50.12 & 73.54 & 77.62 & 55.12 & 72.75 & 78.39 & 60.72 & 71.64 & 77.43 \\
    MoCo-v3 & 73.19 & 76.65 & 78.26 & 55.88 & 69.21 & 76.69 & 50.32 & 63.71 & 71.31 & 51.38 & 71.66 & 74.33 & 62.13 & 76.74 & 75.29 & 54.61 & 74.26 & 77.68 \\
    SimSiam & 73.15 & 72.70 & 75.63 & 62.52 & 69.31 & 76.38 & 55.16 & 62.91 & 71.31 & 49.30 & 69.47 & 75.92 & 58.35 & 72.89 & 75.31 & 58.25 & 68.61 & 77.41 \\
    TS-TCC & 70.73 & 75.88 & 78.91 & 53.54 & 66.98 & 77.87 & 48.04 & 61.79 & 71.18 & 43.34 & 69.48 & 78.23 & 57.07 & 73.62 & 78.72 & 55.26 & 68.48 & 76.79 \\
    CLOCS & 68.94 & 73.36 & 76.31 & 57.94 & 72.55 & 76.24 & 51.97 & 57.96 & 72.65 & 47.19 & 71.88 & 76.31 & 59.59 & 77.78 & 77.49 & 54.38 & 71.93 & 76.13 \\
    ASTCL & 72.51 & 77.31 & 81.02 & 61.86 & 68.77 & 76.51 & 44.14 & 60.93 & 66.99 & 52.38 & 71.98 & 76.05 & 57.90 & 77.01 & 79.51 & 56.40 & 70.87 & 75.79 \\
    CRT & 69.68 & 78.24 & 77.24 & 61.98 & 70.82 & 78.67 & 46.41 & 59.49 & 68.73 & 47.44 & 73.52 & 74.41 & 58.01 & 76.43 & 82.03 & 56.21 & 73.70 & 78.80 \\
    ST-MEM & 61.12 & 66.87 & 71.36 & 54.12 & 57.86 & 63.59 & 55.71 & 59.99 & 66.07 & 51.12 & 65.44 & 74.85 & 56.69 & 63.32 & 70.39 & 59.77 & 66.87 & 71.36 \\
    \midrule
    \multicolumn{19}{l}{\textbf{\textit{Multimodal Methods}}}\\
    \midrule
    MERL (ResNet) & 82.39 & 86.27 & 88.67 & 64.90 & 80.56 & 84.72 & 58.26 & 72.43 & 79.65 & 53.33 & 82.88 & 88.34 & 70.33 & 85.32 & 90.57 & 66.60 & 82.74 & 87.95 \\
    MERL (ViT) & 78.64 & 83.90 & 85.27 & 61.41 & 77.55 & 82.98 & 56.32 & 69.11 & 77.66 & 52.16 & 78.07 & 81.83 & 69.25 & 82.82 & 89.44 & 63.66 & 78.67 & 84.87 \\
    \midrule
    \textbf{K-MERL (Ours)} & \textbf{84.19} & \textbf{87.71} & \textbf{89.83} & \textbf{68.22} & \textbf{81.54} & \textbf{88.00} & \textbf{60.11} & \textbf{73.71} & \textbf{81.48} & \textbf{63.72} & \textbf{84.16} & \textbf{91.04} & \textbf{71.91} & \textbf{86.13} & \textbf{91.26} & \textbf{69.51} & \textbf{83.53} & \textbf{93.71} \\
\bottomrule[1.2pt]
\end{tabular}
}
\label{tab: linear cls}
\vspace{-5pt}
\end{table*}

\noindent\textbf{State-of-the-art on Unseen Disease Prediction.} Additionally, since we extract cardiac-related entities from reports during pre-training, there may be overlap with categories in downstream tasks. This could provide our model with prior knowledge of certain categories, leading to an unfair comparison with MERL \citep{merl}. To address this, we use Med-CPT \citep{jin2023medcpt}, the text encoder, to extract embeddings for all 277 cardiac-related entities and for all category names in the downstream datasets. We compute the similarity between these embeddings, and if the similarity exceeds 0.95, we consider them overlapped. We identify 35 out of 277 extracted cardiac-related entities that overlap with downstream categories, as listed in Tab \ref{tab: overlap}. We label these as `\textit{Seen Classes},’ while the remaining downstream categories are labeled as `\textit{Unseen Classes}.’

The average F1 score are depicted in Fig \ref{fig: seen_unseen_new}(b). K-MERL outperforms MERL in both seen and unseen categories. Notably, both K-MERL and MERL exhibit performance drops on unseen classes compared to seen classes, demonstrating that we successfully detected an overlap of approximately 12.7\% between the extracted cardiac-related entities from MIMIC-ECG and downstream categories, effectively separating the tasks into `seen’ and `unseen’ groups. The results show that K-MERL performs well not only on categories present during pre-training but also on unseen categories, demonstrating its generalizability. 
Since the original MERL \cite{merl} framework relies on manual prompt engineering (PE) at inference time to enhance performance, we also evaluate MERL with customized prompts, as detailed in Sec. \ref{sec:w/o pe}, to provide a comprehensive comparison. Notably, our method outperforms MERL with PE while being entirely independent of prompt engineering.

\subsection{Performance of Linear Probing}
As shown in Tab \ref{tab: linear cls}, K-MERL consistently outperforms multimodal methods, including MERL \citep{merl} with both ResNet and ViT backbones, as well as all eSSL methods across datasets and data ratios. This highlights K-MERL's robust performance and the quality of its learned ECG features, which not only improve multimodal tasks but also significantly enhance single-modality tasks.

\subsection{Performance with Partial Leads Input}

\begin{figure*}[t!]
    \centering
    \includegraphics[width=0.8\textwidth]{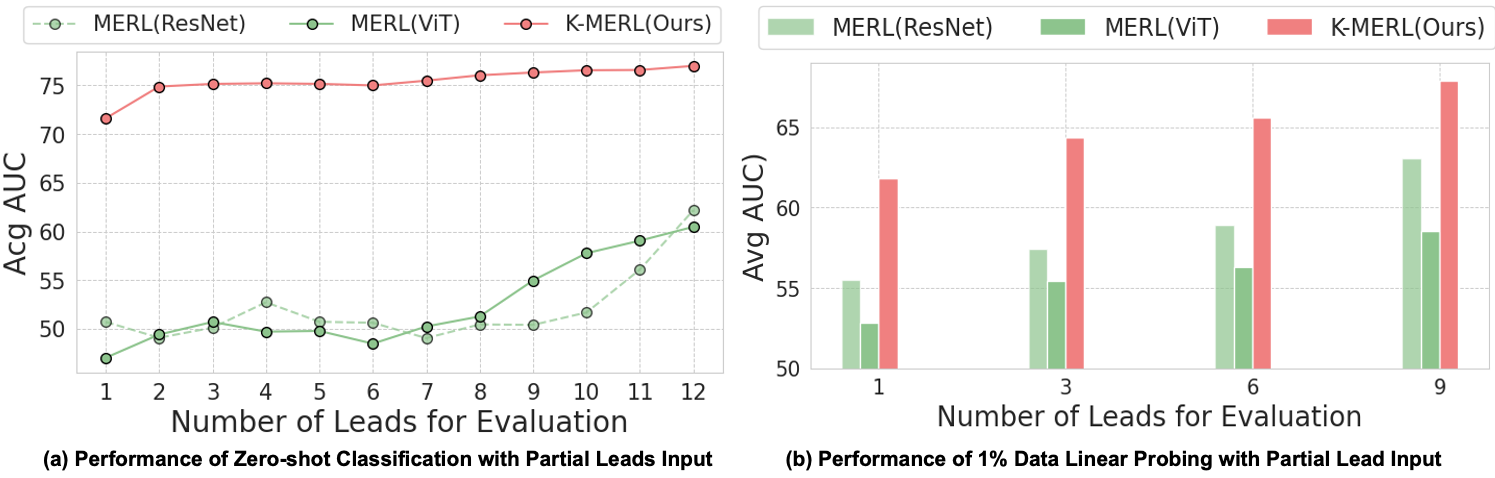}
    \vspace{-10pt}
    \caption{
\small{Performance comparison of K-MERL and MERL with partial lead inputs. \textbf{(a)} Zero-shot classification shows K-MERL consistently outperforming MERL with two backbones across all lead combinations from 1 to 12. \textbf{(b)} Linear probing with 1\% data demonstrates K-MERL's superior performance and robustness, even with limited data and varying lead inputs.}
}
    \label{fig: partial_lead}
    \vspace{-5pt}
\end{figure*}

{As shown in Fig \ref{fig: partial_lead} (a) and (b), K-MERL consistently outperforms MERL across all lead combinations from 1 to 12 in both zero-shot classification and linear probing. Impressively, K-MERL with just a single lead surpasses MERL's performance using all 12 leads. Additionally, K-MERL shows a stable performance trend as the number of leads increases, unlike MERL, which exhibits fluctuations in Fig \ref{fig: partial_lead} (a). This demonstrates the effectiveness of our dynamic lead masking strategy, lead-specific processing, and spatial-temporal positional embeddings, contributing to K-MERL's superior results.}

\section{Analysis}
This section provides extensive ablation studies on the key components of K-MERL and reports zero-shot classification results for single-lead and 12-lead inputs across all downstream datasets. Due to the page limit, we show more ablation studies in Sec \ref{sec: extra ablation}

\begin{figure*}[t!]
    \centering
    \includegraphics[width=0.8\textwidth]{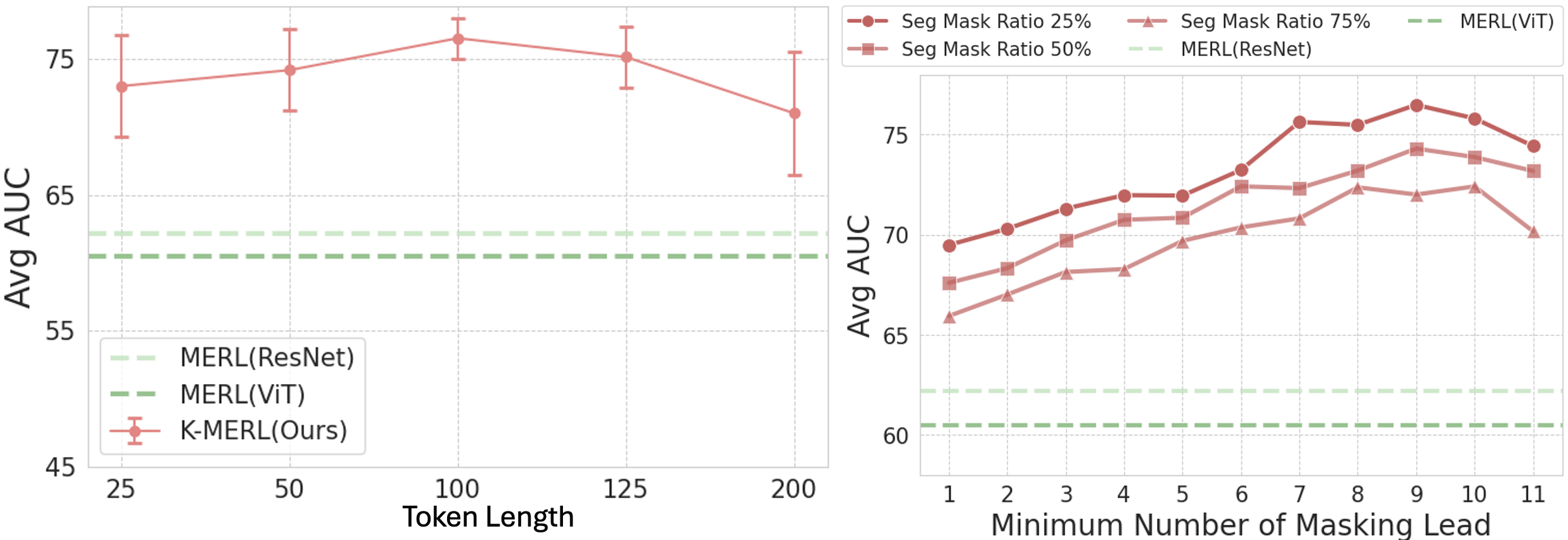}
    \vspace{-5pt}
    \caption{
    \small{Ablation study on zero-shot classification with 12 leads. \textbf{Left:} Performance of K-MERL across varying token lengths, showing optimal results with a token length of 100, consistently outperforming MERL. \textbf{Right:} Impact of different segment masking ratios (25\%, 50\%, 75\%) and the minimum number of masked leads. K-MERL outperforms MERL, with the best performance at a 25\% mask ratio and a minimum of 9 masked leads.}
    }
    \label{fig: abla token mask}
    \vspace{-10pt}
\end{figure*}
%

%
\begin{table*}[t!]
\centering
\caption{\small{Results of various ablation experiments. The best results are \textbf{bolded}.}}%
    \vspace{-10pt}
\subfloat[{Ablating Loss Function.}]{
\resizebox{0.45\linewidth}{!}{
\begin{tabular}{l|cc}
\toprule Loss & 1 Lead & 12 Leads \\
\midrule K-MERL (Ours) & \textbf{71.61} & \textbf{76.52} \\
\midrule \textcolor{red}{\textbf{--}} ECG-Text Alignment ($\mathcal{L}_{\textrm{contrast}}$) & 69.23 & 73.98 \\
\midrule \textcolor{red}{\textbf{--}} ECG-Condition Alignment ($\mathcal{L}_{\textrm{CQ}}$) & 65.44 & 68.95 \\
\bottomrule
\end{tabular}
}\label{tab: abla loss}} 
\hfill
\subfloat[{Effects of Lead-specific Processing.}]{
\resizebox{0.5\linewidth}{!}{
\begin{tabular}{l|cc}
\toprule Methods & 1 Lead & 12 Leads \\
\midrule K-MERL (Ours) & \textbf{71.61} & \textbf{76.52} \\
\midrule \textcolor{red}{\textbf{--}} Lead-specific Tokenization & 68.47 & 74.23 \\
\textcolor{red}{\textbf{--}} Lead-specific Spatial Positional Embedding  & 69.12 & 75.35 \\
\textcolor{red}{\textbf{--}} Lead-agnostic Temporal Positional Embedding  & 70.84 & 75.10 \\
\bottomrule
\end{tabular}
}\label{tab: lead process}} \\
\vspace{2pt}
\subfloat[{Effects of Entities Processing.}]{
\resizebox{0.35\linewidth}{!}{
\begin{tabular}{l|cc}
\toprule Methods & 1 Lead & 12 Leads \\
\midrule K-MERL (Ours) & \textbf{71.61} & \textbf{76.52} \\
\midrule \textcolor{red}{\textbf{--}} Subtype Aggregation & 70.11 & 74.62 \\
\textcolor{red}{\textbf{--}} Merging Duplicated Patterns  & 70.54 & 74.93 \\
\bottomrule
\end{tabular}
}\label{tab: abla condition}}
\hfill
\subfloat[Effects of Masking Strategy.]{
\resizebox{0.33\linewidth}{!}{
\begin{tabular}{l|cc}
\toprule Masking Strategy & 1 Lead & 12 Leads \\
\midrule K-MERL (Ours) & \textbf{71.61} & \textbf{76.52} \\
\midrule \textcolor{red}{\textbf{--}} Lead-independent Segment Masking & 70.32 & 75.21 \\
\textcolor{red}{\textbf{--}} Segment Masking  & 68.93 & 74.74 \\
\textcolor{red}{\textbf{--}} Dynamic Lead Masking & 67.84 & 72.11 \\
\textcolor{red}{\textbf{--}} Lead Masking & 65.41 & 69.10 \\
\bottomrule
\end{tabular}
}\label{tab: abla mask strategy}}
\hfill
\subfloat[Effects of Text Encoder.]{
\resizebox{0.24\linewidth}{!}{
\begin{tabular}{l|cc}
\toprule Text Encoder & 1 Lead & 12 Leads \\
\midrule BioClinicalBERT  & 68.25 & 73.21 \\
\midrule Med-KEBERT & 69.62 & 74.59 \\
\midrule Med-CPT  & \textbf{71.61} & \textbf{76.52} \\ 
\bottomrule
\end{tabular}
}\label{tab: abla textencoder}}
\vspace{-10pt}
\end{table*}

\noindent\textbf{Loss Ablation.}
Tab \ref{tab: abla loss} shows the effect of removing \(\mathcal{L}_{\mathrm{contrast}}\) and \(\mathcal{L}_{\mathrm{CQ}}\) during pre-training. Removing \(\mathcal{L}_{\mathrm{CQ}}\), which excludes structured knowledge from cardiac-related entities, leads to a significant performance drop. While removing \(\mathcal{L}_{\mathrm{contrast}}\) also reduces performance, the impact is less severe. This indicates that both losses are necessary, with cardiac-related entities alignment providing a larger benefit for pre-training.
\\
\noindent\textbf{Tokenization Size.}
In Fig \ref{fig: abla token mask} (a), we ablate the token size \(p\) and find the optimal length to be 100. Larger token sizes (e.g., 200) have a more negative impact than smaller sizes (e.g., 25), likely due to convert multiple segments to one token, which introduces ambiguity. Across all token sizes, K-MERL consistently outperforms MERL \citep{merl}, demonstrating the robustness and effectiveness of our method.
\\
\noindent\textbf{Lead-specific Processing.}
In Tab \ref{tab: lead process}, we ablate the effects of lead-specific tokenization, lead-specific spatial positional embedding, and lead-agnostic temporal embedding. he results show each component enhances K-MERL's performance, with the full combination yielding the best results. The results demonstrate that lead-specific processing is crucial for enabling the ECG multimodal model to recognize lead uniqueness.
\\
\noindent\textbf{Masking Strategy and Ratio.}
Tab \ref{tab: abla mask strategy} shows the results of various masking strategies, where all approaches enhance K-MERL's performance. Removing dynamic lead masking and using a fixed number of masked leads degrades performance, highlighting its importance. Similarly, omitting lead masking during pre-training causes a sharp drop in zero-shot classification, indicating its role in capturing lead-specific features.
Fig \ref{fig: abla token mask} (b) explores mask ratios and lead masking. An optimal configuration is identified with a mask ratio of 25\% and a minimum of 9 masked leads. Increasing the mask ratio beyond this or using more than 9 leads as the minimum for masking leads to a decrease in performance.
\\
\noindent\textbf{Cardiac-related Entities Processing.}
As shown in Tab \ref{tab: abla condition}, both subtype aggregation and merging duplicate entity names improve K-MERL's performance. However, the best results are achieved when both procedures are applied together, indicating they complement each other.
\\
\noindent\textbf{Text Encoder.}
Tab. \ref{tab: abla textencoder} shows Med-CPT \citep{jin2023medcpt} outperforms BioClinicalBERT \citep{alsentzer2019publicly} and Med-KEBERT \citep{kad}, due to contrastive pretraining on a large medical corpus, suggesting contrastive pretraining improves text encoder performance for this task.

\section{Conclusion}
We present K-MERL, a knowledge-enhanced ECG multimodal learning framework capable of processing arbitrary lead inputs. First, we mine cardiac-related entities as structured knowledge from ECG free-text reports using a general LLM, without relying on external domain-specific resources. Next, we align ECG features with these cardiac-related entities to integrate this knowledge into the ECG multimodal learning. Additionally, we introduce lead-specific processing and lead\&segment masking strategies to capture the spatial-temporal patterns unique to each ECG lead, enabling the model to handle varying lead inputs. Our experiments on six downstream ECG classification tasks, along with extensive ablation studies, demonstrate K-MERL's superior zero-shot and linear probing performance compared to existing ECG multimodal and self-supervised learning methods.

\section*{Limitation}
While K-MERL demonstrates promising results in handling arbitrary lead inputs and integrating knowledge from ECG reports, there are some limitations to consider. The framework’s reliance on LLMs for mining cardiac-related entities, though effective, may be limited by the model’s ability to capture highly specialized domain knowledge. Additionally, while our experiments show strong zero-shot and linear probing performance, further evaluation is needed to assess K-MERL's effectiveness in real-world clinical settings, where data quality and noise levels can be more challenging. Future work will focus on enhancing the robustness of knowledge extraction and developing more adaptive strategies for handling diverse ECG data sources.

\clearpage
\bibliography{acl_latex}

\begin{thebibliography}{56}
\providecommand{\natexlab}[1]{#1}

\bibitem[{AI@Meta(2024)}]{llama3modelcard}
AI@Meta. 2024.
\newblock \href {https://github.com/meta-llama/llama3/blob/main/MODEL_CARD.md} {Llama 3 model card}.

\bibitem[{Alizadeh~Meghrazi et~al.(2020)Alizadeh~Meghrazi, Tian, Mahnam, Bhattachan, Eskandarian, Taghizadeh~Kakhki, Popovic, and Lankarany}]{alizadeh2020multichannel}
Milad Alizadeh~Meghrazi, Yupeng Tian, Amin Mahnam, Presish Bhattachan, Ladan Eskandarian, Sara Taghizadeh~Kakhki, Milos~R Popovic, and Milad Lankarany. 2020.
\newblock Multichannel ecg recording from waist using textile sensors.
\newblock \emph{BioMedical Engineering OnLine}, 19:1--18.

\bibitem[{Alsentzer et~al.(2019)Alsentzer, Murphy, Boag, Weng, Jin, Naumann, and McDermott}]{alsentzer2019publicly}
Emily Alsentzer, John~R Murphy, Willie Boag, Wei-Hung Weng, Di~Jin, Tristan Naumann, and Matthew McDermott. 2019.
\newblock Publicly available clinical bert embeddings.
\newblock \emph{arXiv preprint arXiv:1904.03323}.

\bibitem[{Arnaout et~al.(2016)Arnaout, Nah, Marcus, Tseng, Foster, Harris, Divanji, Olgin, Klein, Gonzalez et~al.}]{arnaout2016peripartum}
Rima Arnaout, Gregory Nah, Gregory~M Marcus, Zian~H Tseng, Elyse Foster, Ian Harris, Punag Divanji, Jeffrey Olgin, Liviu Klein, Juan Gonzalez, et~al. 2016.
\newblock Peripartum cardiomyopathy and hypertensive pregnancy subtypes among 1.6 million pregnancies in california independently predict subsequent incident myocardial infarction, heart failure, and stroke.
\newblock \emph{Circulation}, 134(suppl\_1):A14574--A14574.

\bibitem[{Bray et~al.(2021)Bray, Lloyd, Adenwalla, Kelly, Wareham, and Halcox}]{bray2021single}
Jonathan James~Hyett Bray, Elin~Fflur Lloyd, Firdaus Adenwalla, Sarah Kelly, Kathie Wareham, and Julian~PJ Halcox. 2021.
\newblock Single-lead ecgs (alivecor) are a feasible, cost-effective and safer alternative to 12-lead ecgs in community diagnosis and monitoring of atrial fibrillation.
\newblock \emph{BMJ open quality}, 10(1):e001270.

\bibitem[{Brieger et~al.(2000)Brieger, Mak, Miller, Califf, Topol, Investigators et~al.}]{brieger2000hierarchy}
David~B Brieger, Koon-Hou Mak, David~P Miller, Robert~M Califf, Eric~J Topol, GUSTO-I Investigators, et~al. 2000.
\newblock Hierarchy of risk based on history and location of prior myocardial infarction in the thrombolytic era.
\newblock \emph{American Heart Journal}, 140(1):29--33.

\bibitem[{Chamadiya et~al.(2013)Chamadiya, Mankodiya, Wagner, and Hofmann}]{chamadiya2013textile}
Bhavin Chamadiya, Kunal Mankodiya, Manfred Wagner, and Ulrich~G Hofmann. 2013.
\newblock Textile-based, contactless ecg monitoring for non-icu clinical settings.
\newblock \emph{Journal of Ambient Intelligence and Humanized Computing}, 4:791--800.

\bibitem[{Chen et~al.(2020)Chen, Kornblith, Norouzi, and Hinton}]{simclr}
Ting Chen, Simon Kornblith, Mohammad Norouzi, and Geoffrey Hinton. 2020.
\newblock A simple framework for contrastive learning of visual representations.
\newblock In \emph{International conference on machine learning}, pages 1597--1607. PMLR.

\bibitem[{Dai et~al.(2016)Dai, Xiao, Chen, Lin, Wu, and Chen}]{dai2016low}
Ming Dai, Xueliang Xiao, Xin Chen, Haoming Lin, Wanqing Wu, and Siping Chen. 2016.
\newblock A low-power and miniaturized electrocardiograph data collection system with smart textile electrodes for monitoring of cardiac function.
\newblock \emph{Australasian physical \& engineering sciences in medicine}, 39:1029--1040.

\bibitem[{Delbrouck et~al.(2024)Delbrouck, Chambon, Chen, Varma, Johnston, Blankemeier, Van~Veen, Bui, Truong, and Langlotz}]{delbrouck2024radgraph}
Jean-Benoit Delbrouck, Pierre Chambon, Zhihong Chen, Maya Varma, Andrew Johnston, Louis Blankemeier, Dave Van~Veen, Tan Bui, Steven Truong, and Curtis Langlotz. 2024.
\newblock Radgraph-xl: A large-scale expert-annotated dataset for entity and relation extraction from radiology reports.
\newblock In \emph{Findings of the Association for Computational Linguistics ACL 2024}, pages 12902--12915.

\bibitem[{Fontana et~al.(2019)Fontana, Martins, Camenzind, Rossi, Baty, Boesch, Schoch, Brutsche, and Annaheim}]{fontana2019clinical}
Piero Fontana, Neusa R~Ad{\~a}o Martins, Martin Camenzind, Ren{\'e}~M Rossi, Florent Baty, Maximilian Boesch, Otto~D Schoch, Martin~H Brutsche, and Simon Annaheim. 2019.
\newblock Clinical applicability of a textile 1-lead ecg device for overnight monitoring.
\newblock \emph{Sensors}, 19(11):2436.

\bibitem[{Gow et~al.()Gow, Pollard, Nathanson, Johnson, Moody, Fernandes, Greenbaum, Berkowitz, Moukheiber, Eslami et~al.}]{mimicecg}
Brian Gow, Tom Pollard, Larry~A Nathanson, Alistair Johnson, Benjamin Moody, Chrystinne Fernandes, Nathaniel Greenbaum, Seth Berkowitz, Dana Moukheiber, Parastou Eslami, et~al.
\newblock Mimic-iv-ecg-diagnostic electrocardiogram matched subset.

\bibitem[{Huang et~al.(2020)Huang, Fang, Lu, Yan, Yang, and Xu}]{huang2020dual}
Xin Huang, Yu~Fang, Mingming Lu, Fengqi Yan, Jun Yang, and Yilu Xu. 2020.
\newblock Dual-ray net: automatic diagnosis of thoracic diseases using frontal and lateral chest x-rays.
\newblock \emph{Journal of Medical Imaging and Health Informatics}, 10(2):348--355.

\bibitem[{Huang and Yen(2022)}]{SPNv2}
Yu~Huang and Yen. 2022.
\newblock Snippet policy network v2: Knee-guided neuroevolution for multi-lead ecg early classification.
\newblock \emph{IEEE Transactions on Neural Networks and Learning Systems}.

\bibitem[{Jahrsdoerfer et~al.(2005)Jahrsdoerfer, Giuliano, and Stephens}]{jahrsdoerfer2005clinical}
Mary Jahrsdoerfer, Karen Giuliano, and Dean Stephens. 2005.
\newblock Clinical usefulness of the easi 12-lead continuous electrocardiographic monitoring system.
\newblock \emph{Critical care nurse}, 25(5):28--37.

\bibitem[{Jin et~al.(2024)Jin, Wang, Li, Huang, Pan, and Hong}]{jinreading}
Jiarui Jin, Haoyu Wang, Jun Li, Sichao Huang, Jiahui Pan, and Shenda Hong. 2024.
\newblock Reading your heart: Learning ecg words and sentences via pre-training ecg language model.
\newblock In \emph{Artificial Intelligence and Data Science for Healthcare: Bridging Data-Centric AI and People-Centric Healthcare}.

\bibitem[{Jin et~al.(2023)Jin, Kim, Chen, Comeau, Yeganova, Wilbur, and Lu}]{jin2023medcpt}
Qiao Jin, Won Kim, Qingyu Chen, Donald~C Comeau, Lana Yeganova, W~John Wilbur, and Zhiyong Lu. 2023.
\newblock Medcpt: Contrastive pre-trained transformers with large-scale pubmed search logs for zero-shot biomedical information retrieval.
\newblock \emph{Bioinformatics}, 39(11):btad651.

\bibitem[{Junior et~al.(2023)Junior, Reyna, Hong, Gupta, Ghanta, Sameni, Rosand, Aguirre, Qiao, Clifford, and Westover}]{moura2023harvard}
Valdery~Moura Junior, Matthew Reyna, Shenda Hong, Aditya Gupta, Manohar Ghanta, Reza Sameni, Jonathan Rosand, Aaron Aguirre, Li~Qiao, Gari Clifford, and M.~Brandon Westover. 2023.
\newblock \href {https://doi.org/10.60508/g072-7n95} {Harvard-emory ecg database (version 1.0)}.

\bibitem[{King~Jr(2017)}]{king2017approach}
Talmadge~E King~Jr. 2017.
\newblock Approach to the adult with interstitial lung disease: Diagnostic testing.
\newblock \emph{UpToDate. Waltham, MA. Accessed on: March}, 25.

\bibitem[{Kiyasseh et~al.(2021)Kiyasseh, Zhu, and Clifton}]{clocs}
Dani Kiyasseh, Tingting Zhu, and David~A Clifton. 2021.
\newblock Clocs: Contrastive learning of cardiac signals across space, time, and patients.
\newblock In \emph{International Conference on Machine Learning}, pages 5606--5615. PMLR.

\bibitem[{Kotelnik et~al.(2021)Kotelnik, Pesce, Masterton, Marshall, Pigott, Bialek, Winslow, and Maloney}]{kotelnik202112}
Vladimir Kotelnik, Kevin Pesce, William~M Masterton, Robert~T Marshall, Gregson Pigott, Nathaniel Bialek, Jason Winslow, and Lauren~M Maloney. 2021.
\newblock 12-lead electrocardiograms acquired and transmitted by emergency medical technicians are of diagnostic quality and positively impact patient care.
\newblock \emph{Prehospital and Disaster Medicine}, 36(1):47--50.

\bibitem[{Lai et~al.(2023)Lai, Tan, Wang, Ji, Guo, Han, Shi, Feng, and Yang}]{lai2023practical}
Jiewei Lai, Huixin Tan, Jinliang Wang, Lei Ji, Jun Guo, Baoshi Han, Yajun Shi, Qianjin Feng, and Wei Yang. 2023.
\newblock Practical intelligent diagnostic algorithm for wearable 12-lead ecg via self-supervised learning on large-scale dataset.
\newblock \emph{Nature Communications}, 14(1):3741.

\bibitem[{Lalam et~al.(2023)Lalam, Kunderu, Ghosh, A, Awasthi, Prasad, Lopez-Jimenez, Attia, Asirvatham, Friedman, Barve, and Babu}]{lalam2023ecg}
Sravan~Kumar Lalam, Hari~Krishna Kunderu, Shayan Ghosh, Harish~Kumar A, Samir Awasthi, Ashim Prasad, Francisco Lopez-Jimenez, Zachi~I Attia, Samuel Asirvatham, Paul Friedman, Rakesh Barve, and Melwin Babu. 2023.
\newblock \href {https://openreview.net/forum?id=UxmvCwuTMG} {{ECG} representation learning with multi-modal {EHR} data}.
\newblock \emph{Transactions on Machine Learning Research}.

\bibitem[{Li et~al.(2023)Li, Liu, Cheng, Arcucci, and Hong}]{li2023frozen}
Jun Li, Che Liu, Sibo Cheng, Rossella Arcucci, and Shenda Hong. 2023.
\newblock Frozen language model helps ecg zero-shot learning.
\newblock \emph{arXiv preprint arXiv:2303.12311}.

\bibitem[{Li et~al.(2019)Li, Xu, Wang, Jiang, and Liu}]{li2019attention}
Liu Li, Mai Xu, Xiaofei Wang, Lai Jiang, and Hanruo Liu. 2019.
\newblock Attention based glaucoma detection: A large-scale database and cnn model.
\newblock In \emph{Proceedings of the IEEE/CVF conference on computer vision and pattern recognition}, pages 10571--10580.

\bibitem[{Liu et~al.(2023{\natexlab{a}})Liu, Cheng, Ding, and Arcucci}]{scdnn}
Che Liu, Sibo Cheng, Weiping Ding, and Rossella Arcucci. 2023{\natexlab{a}}.
\newblock Spectral cross-domain neural network with soft-adaptive threshold spectral enhancement.
\newblock \emph{arXiv preprint arXiv:2301.10171}.

\bibitem[{Liu et~al.(2023{\natexlab{b}})Liu, Ouyang, Cheng, Shah, Bai, and Arcucci}]{liu2023g2d}
Che Liu, Cheng Ouyang, Sibo Cheng, Anand Shah, Wenjia Bai, and Rossella Arcucci. 2023{\natexlab{b}}.
\newblock G2d: From global to dense radiography representation learning via vision-language pre-training.
\newblock \emph{arXiv preprint arXiv:2312.01522}.

\bibitem[{Liu et~al.(2023{\natexlab{c}})Liu, Wan, Cheng, Zhang, and Arcucci}]{liu2023etp}
Che Liu, Zhongwei Wan, Sibo Cheng, Mi~Zhang, and Rossella Arcucci. 2023{\natexlab{c}}.
\newblock Etp: Learning transferable ecg representations via ecg-text pre-training.
\newblock \emph{arXiv preprint arXiv:2309.07145}.

\bibitem[{Liu et~al.(2024)Liu, Wan, Ouyang, Shah, Bai, and Arcucci}]{merl}
Che Liu, Zhongwei Wan, Cheng Ouyang, Anand Shah, Wenjia Bai, and Rossella Arcucci. 2024.
\newblock Zero-shot ecg classification with multimodal learning and test-time clinical knowledge enhancement.
\newblock \emph{arXiv preprint arXiv:2403.06659}.

\bibitem[{Liu et~al.(2018)Liu, Liu, Zhao, Zhang, Wu, Xu, Liu, Ma, Wei, He et~al.}]{cpsc2018}
Feifei Liu, Chengyu Liu, Lina Zhao, Xiangyu Zhang, Xiaoling Wu, Xiaoyan Xu, Yulin Liu, Caiyun Ma, Shoushui Wei, Zhiqiang He, et~al. 2018.
\newblock An open access database for evaluating the algorithms of electrocardiogram rhythm and morphology abnormality detection.
\newblock \emph{Journal of Medical Imaging and Health Informatics}, 8(7):1368--1373.

\bibitem[{Madias(2003)}]{madias2003comparison}
John~E Madias. 2003.
\newblock A comparison of 2-lead, 6-lead, and 12-lead ecgs in patients with changing edematous states: implications for the employment of quantitative electrocardiography in research and clinical applications.
\newblock \emph{Chest}, 124(6):2057--2063.

\bibitem[{Maheshwari et~al.(2014)Maheshwari, Acharyya, Rajalakshmi, Puddu, and Schiariti}]{maheshwari2014accurate}
Sidharth Maheshwari, Amit Acharyya, Pachamuthu Rajalakshmi, Paolo~Emilio Puddu, and Michele Schiariti. 2014.
\newblock Accurate and reliable 3-lead to 12-lead ecg reconstruction methodology for remote health monitoring applications.
\newblock \emph{IRBM}, 35(6):341--350.

\bibitem[{Na et~al.(2023)Na, Park, Tae, and Joo}]{stmem}
Yeongyeon Na, Minje Park, Yunwon Tae, and Sunghoon Joo. 2023.
\newblock Guiding masked representation learning to capture spatio-temporal relationship of electrocardiogram.
\newblock In \emph{The Twelfth International Conference on Learning Representations}.

\bibitem[{Nonogi et~al.(2008)Nonogi, Yokoyama, Otsuka, Kasahara, Kataoka, Kokubu, and Sase}]{nonogi2008abstract}
Hiroshi Nonogi, Hiroyuki Yokoyama, Yoritaka Otsuka, Yoichiro Kasahara, Yu~Kataoka, Nobuaki Kokubu, and Kazuhiro Sase. 2008.
\newblock Abstract p182: Usefulness of mobile telemedicine system in real-time transmission of out-of-hospital 12-lead ecg.

\bibitem[{Okshina et~al.(2019)Okshina, Lukiyanov, Drapkina, Klyashtorny, Kudryashov, Belova, Deev, Makoveeva, and Boytsov}]{okshina2019p5352}
EY~Okshina, MM~Lukiyanov, OM~Drapkina, VG~Klyashtorny, EV~Kudryashov, EN~Belova, AD~Deev, AN~Makoveeva, and SA~Boytsov. 2019.
\newblock P5352 the main characteristics and multimorbidity in patients with history of stroke, myocardial infarction and their combination (hospital registry data).
\newblock \emph{European Heart Journal}, 40(Supplement\_1):ehz746--0319.

\bibitem[{Pham et~al.(2024)Pham, Saeed, and Ma}]{pham2024c}
Manh Pham, Aaqib Saeed, and Dong Ma. 2024.
\newblock C-melt: Contrastive enhanced masked auto-encoders for ecg-language pre-training.
\newblock \emph{arXiv preprint arXiv:2410.02131}.

\bibitem[{Phan et~al.(2024)Phan, Xie, Qi, Liu, Liu, Zhang, Liao, Wu, To, and Verjans}]{phan2024decomposing}
Vu~Minh~Hieu Phan, Yutong Xie, Yuankai Qi, Lingqiao Liu, Liyang Liu, Bowen Zhang, Zhibin Liao, Qi~Wu, Minh-Son To, and Johan~W Verjans. 2024.
\newblock Decomposing disease descriptions for enhanced pathology detection: A multi-aspect vision-language pre-training framework.
\newblock In \emph{Proceedings of the IEEE/CVF Conference on Computer Vision and Pattern Recognition}, pages 11492--11501.

\bibitem[{Quinn et~al.(2020)Quinn, Watkins, Hampton, Halter, Weston, Gale, Gavalova, Driscoll, Davies, and Snooks}]{quinn2020has}
T~Quinn, A~Watkins, C~Hampton, M~Halter, C~Weston, CP~Gale, L~Gavalova, T~Driscoll, G~Davies, and HA~Snooks. 2020.
\newblock Has the proportion of patients diagnosed with myocardial infarction that receives a 12 ecg in the prehospital setting in the uk changed over time?
\newblock \emph{European Heart Journal}, 41(Supplement\_2):ehaa946--1650.

\bibitem[{Ribeiro et~al.(2020)Ribeiro, Ribeiro, Paix{\~a}o, Oliveira, Gomes, Canazart, Ferreira, Andersson, Macfarlane, Meira~Jr et~al.}]{ribeiro2020automatic}
Ant{\^o}nio~H Ribeiro, Manoel~Horta Ribeiro, Gabriela~MM Paix{\~a}o, Derick~M Oliveira, Paulo~R Gomes, J{\'e}ssica~A Canazart, Milton~PS Ferreira, Carl~R Andersson, Peter~W Macfarlane, Wagner Meira~Jr, et~al. 2020.
\newblock Automatic diagnosis of the 12-lead ecg using a deep neural network.
\newblock \emph{Nature communications}, 11(1):1760.

\bibitem[{Sangha et~al.(2024)Sangha, Khunte, Holste, Mortazavi, Wang, Oikonomou, and Khera}]{sangha2024biometric}
Veer Sangha, Akshay Khunte, Gregory Holste, Bobak~J Mortazavi, Zhangyang Wang, Evangelos~K Oikonomou, and Rohan Khera. 2024.
\newblock Biometric contrastive learning for data-efficient deep learning from electrocardiographic images.
\newblock \emph{Journal of the American Medical Informatics Association}, 31(4):855--865.

\bibitem[{Sawano et~al.(2022)Sawano, Kodera, Takeuchi, Sukeda, Katsushika, and Komuro}]{sawano2022masked}
Shinnosuke Sawano, Satoshi Kodera, Hirotoshi Takeuchi, Issei Sukeda, Susumu Katsushika, and Issei Komuro. 2022.
\newblock Masked autoencoder-based self-supervised learning for electrocardiograms to detect left ventricular systolic dysfunction.
\newblock In \emph{NeurIPS 2022 Workshop on Learning from Time Series for Health}.

\bibitem[{Swor et~al.(2006)Swor, Hegerberg, McHugh-McNally, Goldstein, and McEachin}]{swor2006prehospital}
Robert Swor, Stacey Hegerberg, Ann McHugh-McNally, Mark Goldstein, and Christine~C McEachin. 2006.
\newblock Prehospital 12-lead ecg: efficacy or effectiveness?
\newblock \emph{Prehospital Emergency Care}, 10(3):374--377.

\bibitem[{Thygesen et~al.(2018)Thygesen, Alpert, Jaffe, Chaitman, Bax, Morrow, White, and on~behalf of the Joint European Society of Cardiology (ESC)/American College of Cardiology (ACC)/American Heart Association (AHA)/World Heart Federation (WHF) Task Force for the Universal Definition~of Myocardial~Infarction}]{thygesen2018fourth}
Kristian Thygesen, Joseph~S Alpert, Allan~S Jaffe, Bernard~R Chaitman, Jeroen~J Bax, David~A Morrow, Harvey~D White, and Executive~Group on~behalf of the Joint European Society of Cardiology (ESC)/American College of Cardiology (ACC)/American Heart Association (AHA)/World Heart Federation (WHF) Task Force for the Universal Definition~of Myocardial~Infarction. 2018.
\newblock Fourth universal definition of myocardial infarction (2018).
\newblock \emph{Circulation}, 138(20):e618--e651.

\bibitem[{Wagner et~al.(2020)Wagner, Strodthoff, Bousseljot, Kreiseler, Lunze, Samek, and Schaeffter}]{wagner2020ptb}
Patrick Wagner, Nils Strodthoff, Ralf-Dieter Bousseljot, Dieter Kreiseler, Fatima~I Lunze, Wojciech Samek, and Tobias Schaeffter. 2020.
\newblock Ptb-xl, a large publicly available electrocardiography dataset.
\newblock \emph{Scientific data}, 7(1):1--15.

\bibitem[{Wan et~al.(2024)Wan, Liu, Wang, Tao, Shen, Peng, Fu, Arcucci, Yao, and Zhang}]{wan2024electrocardiogram}
Zhongwei Wan, Che Liu, Xin Wang, Chaofan Tao, Hui Shen, Zhenwu Peng, Jie Fu, Rossella Arcucci, Huaxiu Yao, and Mi~Zhang. 2024.
\newblock Electrocardiogram instruction tuning for report generation.
\newblock \emph{arXiv preprint arXiv:2403.04945}.

\bibitem[{Wan et~al.(2023)Wan, Liu, Zhang, Fu, Wang, Cheng, Ma, Quilodr{\'a}n-Casas, and Arcucci}]{wan2023med}
Zhongwei Wan, Che Liu, Mi~Zhang, Jie Fu, Benyou Wang, Sibo Cheng, Lei Ma, C{\'e}sar Quilodr{\'a}n-Casas, and Rossella Arcucci. 2023.
\newblock Med-unic: Unifying cross-lingual medical vision-language pre-training by diminishing bias.
\newblock \emph{arXiv preprint arXiv:2305.19894}.

\bibitem[{Wang et~al.(2023)Wang, Feng, Ge, Zhou, Zhou, and Wang}]{astcl}
Ning Wang, Panpan Feng, Zhaoyang Ge, Yanjie Zhou, Bing Zhou, and Zongmin Wang. 2023.
\newblock Adversarial spatiotemporal contrastive learning for electrocardiogram signals.
\newblock \emph{IEEE Transactions on Neural Networks and Learning Systems}.

\bibitem[{Wu et~al.(2023)Wu, Zhang, Zhang, Wang, and Xie}]{wu2023medklip}
Chaoyi Wu, Xiaoman Zhang, Ya~Zhang, Yanfeng Wang, and Weidi Xie. 2023.
\newblock Medklip: Medical knowledge enhanced language-image pre-training in radiology.
\newblock \emph{arXiv preprint arXiv:2301.02228}.

\bibitem[{Wu et~al.(2021)Wu, Agu, Lourentzou, Sharma, Paguio, Yao, Dee, Mitchell, Kashyap, Giovannini et~al.}]{wu2021chest}
Joy Wu, Nkechinyere Agu, Ismini Lourentzou, Arjun Sharma, Joseph Paguio, Jasper~Seth Yao, Edward~Christopher Dee, William Mitchell, Satyananda Kashyap, Andrea Giovannini, et~al. 2021.
\newblock Chest imagenome dataset.
\newblock \emph{Physio Net}.

\bibitem[{Yu et~al.(2024)Yu, Guo, and Sano}]{yu2024ecg}
Han Yu, Peikun Guo, and Akane Sano. 2024.
\newblock Ecg semantic integrator (esi): A foundation ecg model pretrained with llm-enhanced cardiological text.
\newblock \emph{arXiv preprint arXiv:2405.19366}.

\bibitem[{Zhang et~al.(2022)Zhang, Liu, Shi, Chang, Wang, He, and Huang}]{zhang2022maefe}
Huaicheng Zhang, Wenhan Liu, Jiguang Shi, Sheng Chang, Hao Wang, Jin He, and Qijun Huang. 2022.
\newblock Maefe: Masked autoencoders family of electrocardiogram for self-supervised pretraining and transfer learning.
\newblock \emph{IEEE Transactions on Instrumentation and Measurement}, 72:1--15.

\bibitem[{Zhang and Frick(2019)}]{zhang2019all}
Qingxue Zhang and Kyle Frick. 2019.
\newblock All-ecg: A least-number of leads ecg monitor for standard 12-lead ecg tracking during motion.
\newblock In \emph{2019 IEEE Healthcare Innovations and Point of Care Technologies,(HI-POCT)}, pages 103--106. IEEE.

\bibitem[{Zhang et~al.(2023)Zhang, Wu, Zhang, Xie, and Wang}]{kad}
Xiaoman Zhang, Chaoyi Wu, Ya~Zhang, Weidi Xie, and Yanfeng Wang. 2023.
\newblock Knowledge-enhanced visual-language pre-training on chest radiology images.
\newblock \emph{Nature Communications}, 14(1):4542.

\bibitem[{Zhao et~al.(2024)Zhao, Zhang, Wang, Han, Chen, Huang, Jin, and Kang}]{zhao2024ecg}
Yubao Zhao, Tian Zhang, Xu~Wang, Puyu Han, Tong Chen, Linlin Huang, Youzhu Jin, and Jiaju Kang. 2024.
\newblock Ecg-chat: A large ecg-language model for cardiac disease diagnosis.
\newblock \emph{arXiv preprint arXiv:2408.08849}.

\bibitem[{Zheng et~al.(2022)Zheng, Guo, and Chu}]{csn1}
J~Zheng, H~Guo, and H~Chu. 2022.
\newblock A large scale 12-lead electrocardiogram database for arrhythmia study (version 1.0. 0).
\newblock \emph{PhysioNet 2022Available online: http://physionet. org/content/ecg-arrhythmia/1.0. 0/(accessed on 23 November 2022)}.

\bibitem[{Zheng et~al.(2020)Zheng, Chu, Struppa, Zhang, Yacoub, El-Askary, Chang, Ehwerhemuepha, Abudayyeh, Barrett et~al.}]{csn2}
Jianwei Zheng, Huimin Chu, Daniele Struppa, Jianming Zhang, Sir~Magdi Yacoub, Hesham El-Askary, Anthony Chang, Louis Ehwerhemuepha, Islam Abudayyeh, Alexander Barrett, et~al. 2020.
\newblock Optimal multi-stage arrhythmia classification approach.
\newblock \emph{Scientific reports}, 10(1):2898.

\end{thebibliography}
\clearpage

\appendix


\section{Related Work}
\label{sec: realted}

\subsection{ECG Representation Learning}
Recently, ECG self-supervised learning (eSSL) has shown promise in learning ECG representations from unannotated signals~\citep{lai2023practical, simclr, sangha2024biometric}. Contrastive methods such as CLOCS~\citep{clocs} and ASTCL~\citep{astcl} explore temporal and spatial invariance, while generative techniques~\citep{zhang2022maefe, sawano2022masked, stmem, jinreading} focus on masked segment reconstruction. However, both approaches often lack clinical domain knowledge and are limited to single-modality settings, restricting the quality of learned representations. 

Multimodal learning has shown success in multiple biomedical applications~\citep{wan2023med,liu2023g2d,wu2023medklip}. However, ECG signals pose unique challenges due to their complex spatial-temporal structure, necessitating well-tailored modeling. As a result, few studies have explored multimodal ECG learning. \citep{lalam2023ecg,yu2024ecg} demonstrated the effectiveness of combining ECG and EHR data using large language models (LLMs) to rewrite textual reports. However, their work is restricted to private datasets, making reproducing and comparisons challenging. Other works such as \citep{li2023frozen,liu2023etp} explored multimodal ECG learning for zero-shot classification. However, their methods were over simplistic: They align signals with text without sufficiently capturing the distinctiveness of individual ECG leads, and rely on naive category names as prompts, which fail to capture relative patterns, leading to suboptimal performance. Their limited evaluations on small datasets also fall short of fully assessing multimodal ECG learning in real-world scenarios.
Additionally, works such as \citep{zhao2024ecg,wan2024electrocardiogram} focus on ECG-to-text generation tasks, but their results are not publicly accessible, making reproducing and comparisons difficult.

MERL~\citep{merl} is the first open-source study to demonstrate the potential of ECG multimodal learning in zero-shot classification and linear probing across diverse datasets. Therefore, we mainly compare our work to MERL. However, like other methods, MERL relies on all 12 ECG leads as input and cannot handle arbitrary lead combinations, limiting its applicability in real-world clinical scenarios where all 12 leads may not always be available \citep{jahrsdoerfer2005clinical,madias2003comparison,fontana2019clinical,maheshwari2014accurate}

\subsection{Knowledge Enhanced Medical Multimodal Learning}
Leveraging medical knowledge to improve medical multimodal learning has advanced significantly, particularly in the radiograph domain, with methods like MedKLIP, KAD, and MAVL \citep{kad,wu2023medklip,phan2024decomposing}. These approaches focus on extracting structured knowledge, such as clinical entities from free-text radiology reports, and using this information as an additional supervisory signal to guide multimodal learning. Many models mimic radiological practices or modify structures based on diagnostic routines \citep{li2019attention,huang2020dual,kad,wu2023medklip}. However, they rely heavily on well-annotated knowledge graphs, such as RadGraph \citep{delbrouck2024radgraph} and Chest ImaGenome \citep{wu2021chest}, which require substantial human annotation and are limited to the radiology domain.
Due to the distinct nature of ECG signals compared to radiographs, the above pipelines cannot be directly adapted for ECG multimodal learning. 
Furthermore, CVD has a clear hierarchical structure because conditions can have multiple subtypes, such as myocardial infarction, which can be further classified as inferior or anterior myocardial infarction \citep{thygesen2018fourth}. Unlike lung diseases, typically categorized by morphological or pathological patterns rather than distinct region based subtypes \citep{king2017approach}, directly using only the entity from an ECG report can lead to information loss by ignoring the superclass or subtypes.

\subsection{Challenge in Partial Leads ECG Input}
Currently, full 12 leads ECG data dominates publicly accessible ECG datasets \citep{mimicecg,ribeiro2020automatic,moura2023harvard}. However, in real clinical scenarios, obtaining a standard 12 leads ECG can be excessive and often requires advanced clinical knowledge, which may not always be readily available \citep{chamadiya2013textile,alizadeh2020multichannel,dai2016low}. This makes partial-lead ECG data both crucial and common for practical applications. Despite its importance, partial leads issue is often overlooked and remain unaddressed in existing ECG multimodal representation learning studies. 
To handle partial lead inputs across various downstream tasks, in this work, we design lead-specific processing and dynamic lead masking strategies that enable our model to accept any combination of ECG leads as input.  adaptable to various clinical scenarios \citep{jahrsdoerfer2005clinical,madias2003comparison,fontana2019clinical,maheshwari2014accurate}. We evaluate our model on extensive downstream tasks with partial lead inputs, demonstrating its ability to recognize and adapt to the lead-specific nature of ECG signals.

\section{Pre-training Configuration}
\label{sec: pre config}
Following MERL \citep{merl}, we employ the AdamW optimizer with a learning rate of $2\times10^{-4}$ and a weight decay of $1\times10^{-5}$. Pre-training runs for 50 epochs, with a cosine annealing scheduler for learning rate adjustments. We use a batch size of 512 per GPU, with all experiments conducted on eight NVIDIA A100-80GB GPUs.

\section{Downstream Task Details}

\subsection{Downstream Task Data Split}
We detail the data splits in Tab. \ref{tab:split}. For all datasets, we follow the splits provided by MERL\footnote{https://github.com/cheliu-computation/MERL-ICML2024/tree/main/finetune/data\_split}. The preprocessing for all datasets is also done using MERL's official codebase\footnote{https://github.com/cheliu-computation/MERL-ICML2024/tree/main/finetune}.

\begin{table*}[ht!]
\centering
\caption{Details on Data Split.}
\scalebox{0.8}{
\begin{tabular}{lcrrr}
\toprule[1.2pt]
Dataset & Number of Categories & Train & Valid & Test \\ 
\midrule[1.2pt]
PTBXL-Super \citep{wagner2020ptb} & 5 & 17,084 & 2,146 & 2,158 \\
PTBXL-Sub \citep{wagner2020ptb} & 23 & 17,084 & 2,146 & 2,158 \\
PTBXL-Form \citep{wagner2020ptb} & 19 & 7,197 & 901 & 880 \\
PTBXL-Rhythm \citep{wagner2020ptb} & 12 & 16,832 & 2,100 & 2,098 \\
\midrule
CPSC2018 \citep{cpsc2018} & 9 & 4,950 & 551 & 1,376 \\
CSN \citep{csn1,csn2} & 38 & 16,546 & 1,860 & 4,620 \\
 \bottomrule[1.2pt]
\end{tabular}
}
\label{tab:split}
\end{table*}

\subsection{Downstream Task Configuration}
We detail the key hyperparameters used across all downstream tasks in Tab. \ref{tab: hyper}. For each dataset (PTBXL-Super, PTBXL-Sub, PTBXL-Form, PTBXL-Rhythm, CPSC2018, and CSN), we maintain consistency in the learning rate, batch size, number of epochs, and optimizer configuration with MERL \citep{merl}.

\begin{table*}[ht!]
    \centering
    \caption{Hyperparameter settings on downstream tasks.}
    \scalebox{0.7}{
    \begin{tabular}{ccccccc}
    \toprule[1.2pt]
    & PTBXL-Super & PTBXL-Sub & PTBXL-Form & PTBXL-Rhythm & CPSC2018 & CSN\\
    \midrule[1.2pt]
    Learning rate & 0.001 & 0.001 & 0.001 & 0.001 & 0.001 & 0.001\\
    Batch size & 16 & 16 & 16 & 16 & 16 & 16 \\
    Epochs & 100 & 100 & 100 & 100 & 100 & 100 \\
    Optimizer & AdamW & AdamW & AdamW & AdamW & AdamW & AdamW \\
    Learing rate scheduler & Cosine anealing & Cosine anealing & Cosine anealing & Cosine anealing & Cosine anealing & Cosine anealing \\
    Warump steps & 5 & 5 & 5 & 5 & 5 & 5 \\
    \bottomrule[1.2pt]
    \end{tabular}
    }
    \label{tab: hyper}
\end{table*}

\subsection{Downstream Tasks Implementation}
\label{sec: downstream implment}
\noindent\textbf{Zero-shot Classification.}  
For zero-shot classification, we freeze the entire model and use the original category names from the dataset as entity queries \(\mathcal{Q}\) for input to the cardiac query network, \(\mathcal{F}_{\mathrm{CQ}}\). The ECG signals are converted into ECG feature with $\mathcal{F}_{\mathrm{E}}$, serving as the key and value inputs for \(\mathcal{F}_{\mathrm{CQ}}\). The output of \(\mathcal{F}_{\mathrm{CQ}}\) provides the predicted probabilities for each category.
\\
\noindent\textbf{Linear Probing.}  
For linear probing, we keep the ECG encoder \(\mathcal{F}_{\mathrm{E}}\) frozen and only update the parameters of a randomly initialized linear classifier. We conduct linear probing with \{1\%, 10\%, 100\%\} of the training data. This configuration is used consistently across all linear probing tasks. Further implementation details are provided in the Tab \ref{tab: hyper}.
\\
\noindent\textbf{Partial Lead Setting.}  
In the partial lead setting, we follow the lead order from the MIMIC-ECG dataset \citep{mimicecg}: [I, II, III, aVF, aVR, aVL, V1, V2, V3, V4, V5, V6], progressively expanding the input from a single lead to all 12 leads in sequence. In contrast, since MERL \citep{merl} requires a full 12-lead input, we pad the missing leads with zeros to maintain the 12-lead format.

\subsection{Overlapped Categories}
As described in Sec \ref{sec: zeroshot cls} and Fig \ref{fig: seen_unseen_new}, we observe that 35 categories are present in both the pre-training and downstream datasets, and we list all the class names in Tab \ref{tab: overlap}.

\begin{table*}[ht!]
\centering
\caption{Overlap of cardiac-related entities between downstream tasks and the pretraining dataset.}

\begin{tabular}{|l|l|}
\hline
prolonged qt interval & normal \\ \hline
arrhythmia & first degree av block \\ \hline
anterior myocardial infarction & ventricular premature complex \\ \hline
conduction disturbance & second degree av block \\ \hline
hypertrophy & st depression \\ \hline
atrial premature complex & prolonged pr interval \\ \hline
t wave abnormalities & premature complex \\ \hline
atrial fibrillation & sinus tachycardia \\ \hline
sinus arrhythmia & sinus bradycardia \\ \hline
atrial flutter & supraventricular tachycardia \\ \hline
atrial premature complex & abnormal q wave \\ \hline
av block & left bundle branch block \\ \hline
myocardial infarction & right bundle branch block \\ \hline
st elevation & st-t changes \\ \hline
t wave changes & ventricular bigeminy \\ \hline
ventricular premature complex & sinus tachycardia \\ \hline
atrial flutter & supraventricular tachycardia \\ \hline
atrial tachycardia & \\ \hline
\end{tabular}
\label{tab: overlap}
\end{table*}

\section{State-of-the-Art Without Prompt Engineering}
\label{sec:w/o pe}
\begin{figure}[ht]
    \centering
    \includegraphics[width=0.9\linewidth]{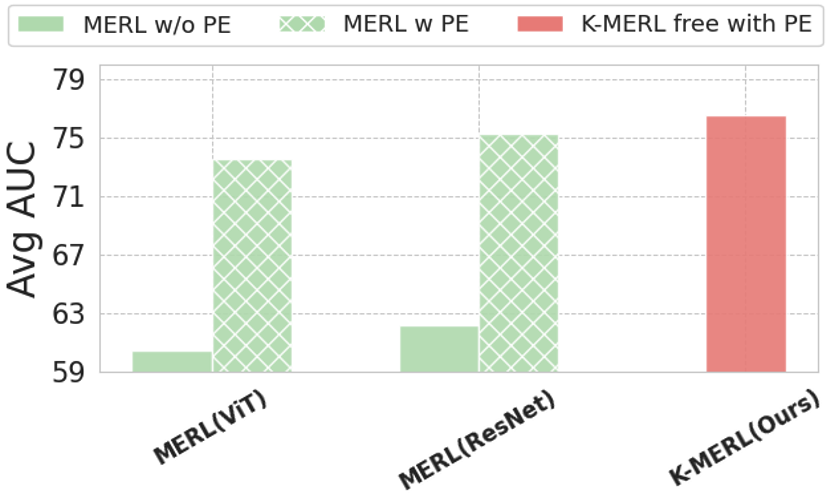}
    \caption{
    \small{Comparison of K-MERL and MERL with prompt engineering (PE). Notably, even though MERL with PE uses customized disease prompts with human effort, K-MERL, \textbf{free with PE}, still surpasses both versions of MERL, demonstrating its generalizability and effectiveness.}
    }
    \label{fig:pe}
\end{figure}

It is important to note that MERL heavily relies on prompt engineering (PE), which requires tailoring the text prompt of each possible disease at \textit{inference time}, querying external knowledge bases using LLM, which is inefficient \citep{merl}.
To fully showcase the our method's capabilities, we compare K-MERL with the PE-enhanced version of MERL in Fig \ref{fig:pe}. Unlike MERL, K-MERL does not depend on any customized disease prompts at inference time, as it has better leveraged cardiac knowledge contained in the reports during pre-training. Despite being free from PE, K-MERL still surpasses MERL with PE, demonstrating the superiority of our approach.

\begin{figure}[ht!]
    \centering
    \parbox[h]{.5\textwidth}{
    \includegraphics[width=0.99\linewidth]{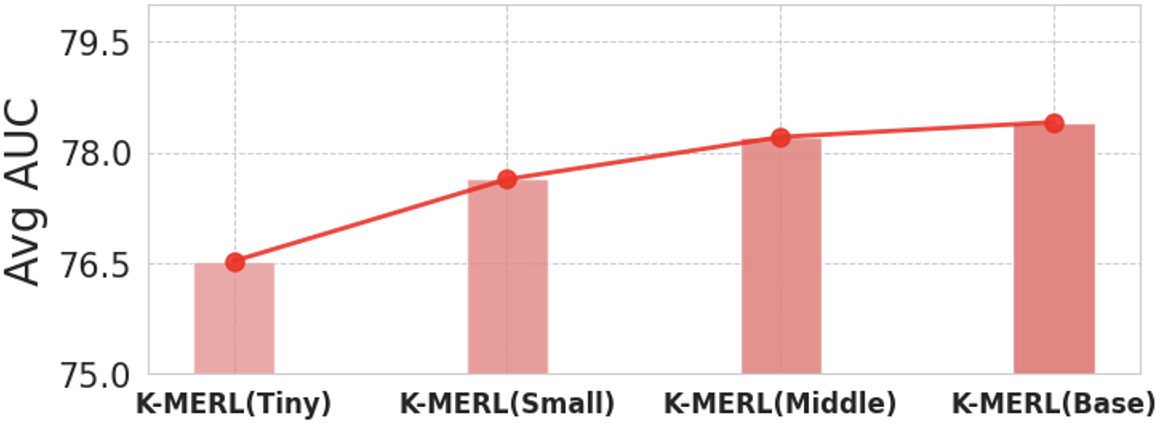}
    }
    \hspace{\fill}
    \parbox[h]{.47\textwidth}{
    \caption{
    \small{Reported performance of zero-shot classification with scaled ECG encoders. As the model size increases from K-MERL(Tiny) to K-MERL(Base), the performance improves, demonstrating the scalability of the model.
    }
    }
    \label{fig:scale}
    }
    \vspace{-20pt}
\end{figure}
\section{Scalability}
\label{sec:scalbility}
We scale our ECG encoder using ViT-Tiny, ViT-Small, ViT-Middle, and ViT-Base, as shown in Fig. \ref{fig:scale}. K-MERL consistently improves as model size increases, demonstrating its scalability for ECG multimodal learning.

\section{Additional Ablation Studies}
\label{sec: extra ablation}

\begin{table*}[ht!]
    \centering
    \caption{Additional Ablation Studies.}
    \subfloat[Effects of LLM on Processing Cardiac-related Entities.\label{tab:abla_llm}]{
    \resizebox{0.35\linewidth}{!}{
    \begin{tabular}{l|cc}
    \toprule[1.2pt]
      Methods & 1 Lead & 12 Leads \\
    \midrule[1.2pt]
    Llama3.1-8B-Instruct  & 68.52 & 74.19 \\
    Gemma-2-9B & 68.94 & 74.47 \\
    Gemma-2-27B & 70.54 & 75.81 \\
    Llama3.1-70B-Instruct  & \textbf{71.61} & \textbf{76.52} \\
    \bottomrule[1.2pt]
    \end{tabular}
    }}
    \hfill
    \subfloat[Effects of the Number of Transformer Layers in the Cardiac Query Network $\mathcal{F}_{\mathrm{CQ}}$\label{tab:abla_pqb_layer}]{
    \resizebox{0.25\linewidth}{!}{
    \begin{tabular}{c|cc}
    \toprule[1.2pt]
      Num of Layers & 1 Lead & 12 Leads \\
    \midrule[1.2pt]
    1  & 69.92 & 72.96 \\
    2  & 70.14 & 73.13 \\
    3  & 70.31 & 74.40 \\
    4  & \textbf{71.61} & \textbf{76.52} \\
    5  & 69.25 & 74.94 \\
    \bottomrule[1.2pt]
    \end{tabular}
    }}
    \hfill
    \subfloat[Effects of the Number of Heads in the Cardiac Query Network $\mathcal{F}_{\mathrm{CQ}}$.\label{tab:abla_pqb_head}]{
    \resizebox{0.25\linewidth}{!}{
    \begin{tabular}{c|cc}
    \toprule[1.2pt]
      Num of Heads & 1 Lead & 12 Leads \\
    \midrule[1.2pt]
    1  & 68.76 & 74.89 \\
    2  & 70.25 & 74.23 \\
    3  & 70.27 & 75.36 \\
    4  & \textbf{71.61} & \textbf{76.52} \\
    5  & 71.23 & 75.48 \\
    \bottomrule[1.2pt]
    \end{tabular}
    }}
\end{table*}
Tab \ref{tab:abla_llm}, \ref{tab:abla_pqb_layer}, and \ref{tab:abla_pqb_head} present the results of additional ablation studies. (1) Tab \ref{tab:abla_llm} shows the impact of various LLMs on processing cardiac-related entities, with Llama3.1-70B-Instruct achieving the best performance across both 1-lead and 12-lead settings. The performance increases with larger LLMs, suggesting that larger models improve cardiac-related entities extraction. (2) Tab \ref{tab:abla_pqb_layer} explores the effects of different numbers of transformer layers in the Cardiac Query Network \(\mathcal{F}_{\mathrm{CQ}}\), showing that performance improves as the number of layers increases and saturates at 4 layers. (3) Tab \ref{tab:abla_pqb_head} examines the effect of the number of attention heads in \(\mathcal{F}_{\mathrm{CQ}}\), with 4 heads providing the best performance.

\end{document}